\documentclass[letterpaper, 10 pt, conference]{ieeeconf}  

\IEEEoverridecommandlockouts                              

\usepackage{graphics} 
\usepackage[demo]{graphicx}
\usepackage{epsfig} 
\usepackage{times} 
\usepackage{amsmath} 
\usepackage{amssymb}  
\usepackage{hyperref}
\usepackage{algorithm}
\usepackage{algorithmic}
\usepackage{textcomp}
\usepackage{float}
\usepackage{algorithm}
\usepackage{algorithmic}
\usepackage{mathtools}
\usepackage{commath}
\usepackage{fixmath}
\usepackage{subcaption}
\usepackage{siunitx}
\usepackage{import}
\usepackage{bbm}
\usepackage[todonotes,table,dvipsnames]{xcolor}
\usepackage[strings]{underscore}
\usepackage{setspace}
\let\Algorithm\algorithm
\renewcommand\algorithm[1][]{\Algorithm[#1]\setstretch{1.15}}
\usepackage{textcomp}

\definecolor{CustomGray}{gray}{0.85}

\usepackage{color}

\def \hrsup_size{0.48}
\def \framesu_size{0.48}

\arrayrulecolor{white}
\usepackage{ntheorem}\newtheorem*{proposition}{Proposition}

\title{\LARGE \bf
Trajectory Advancement \\ during Human-Robot Collaboration
}

\author{Yeshasvi Tirupachuri$^{1}$$^{2}$, Gabriele Nava$^{1}$$^{2}$, Lorenzo Rapetti$^{1}$$^{3}$, Claudia Latella$^{1}$, Daniele Pucci$^{1}$
\thanks{*This work is supported by \href{http://itn-pace.eu/}{PACE} project, Marie Skłodowska-Curie grant agreement No. 642961 and \href{https://andy-project.eu/}{An.Dy} project which has received funding from the European Union\textquotesingle s Horizon 2020 Research and Innovation Programme under grant agreement No. 731540.}
\thanks{$^{1}$Dynamic Interaction Control, Istituto Italiano di Tecnologia, Genova, Italy {\tt\small name.surname@iit.it}}%
\thanks{$^{2}$DIBRIS, University of Genova, Genova, Italy}
\thanks{$^{3}$School of Computer Science, University of Manchester, Manchester, United Kingdom}
}

\begin{document}

\maketitle
\thispagestyle{empty}
\pagestyle{empty}

\begin{abstract}

As technology advances, the barriers between the co-existence of humans and robots are slowly coming down. The prominence of physical interactions for collaboration and cooperation between humans and robots will be an undeniable fact. Rather than exhibiting simple reactive behaviors to human interactions, it is desirable to endow robots with augmented capabilities of exploiting human interactions for successful task completion. Towards that goal, in this paper, we propose a trajectory advancement approach in which we mathematically derive the conditions that facilitate advancing along a reference trajectory by leveraging assistance from helpful interaction wrench present during human-robot collaboration. We validate our approach through experiments conducted with the iCub humanoid robot both in simulation and on the real robot. 

\end{abstract}

\section{INTRODUCTION}
\label{introduction}

The world as we know it is dynamic and evolving. Humans are quite agile in incorporating the latest technologies and augment their \emph{umwelt} effortlessly. It is indisputable that technological progress is headed in the direction of co-existence between humans and robots. The research field of physical Human-Robot Interaction (pHRI) plays a vital role in investigating several aspects that ensure safe co-existence. Robots are evolving to be active agents to support humans in various endeavors and to further augment their capabilities. Human-Robot Collaboration (HRC) has many potential applications such as collaborative manufacturing and elderly assistance.



\begin{figure}[!hbt]
    \centering
    \begin{subfigure}{0.24\textwidth}
        \centering
        \includegraphics[clip, trim=0cm 0cm 0cm 0cm, scale=0.25]{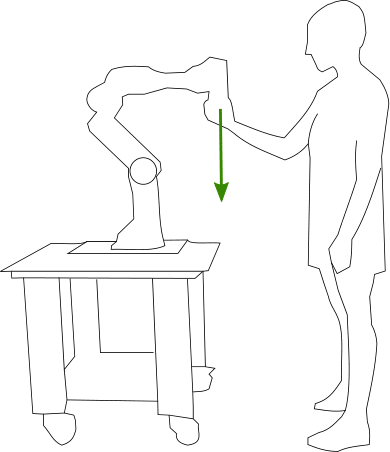}
        \caption{}
        \label{fig:example-cobot-help}
    \end{subfigure}%
    \begin{subfigure}{0.24\textwidth}
        \centering
        \includegraphics[clip, trim=16cm 7.5cm 11.5cm 10cm, scale=0.095]{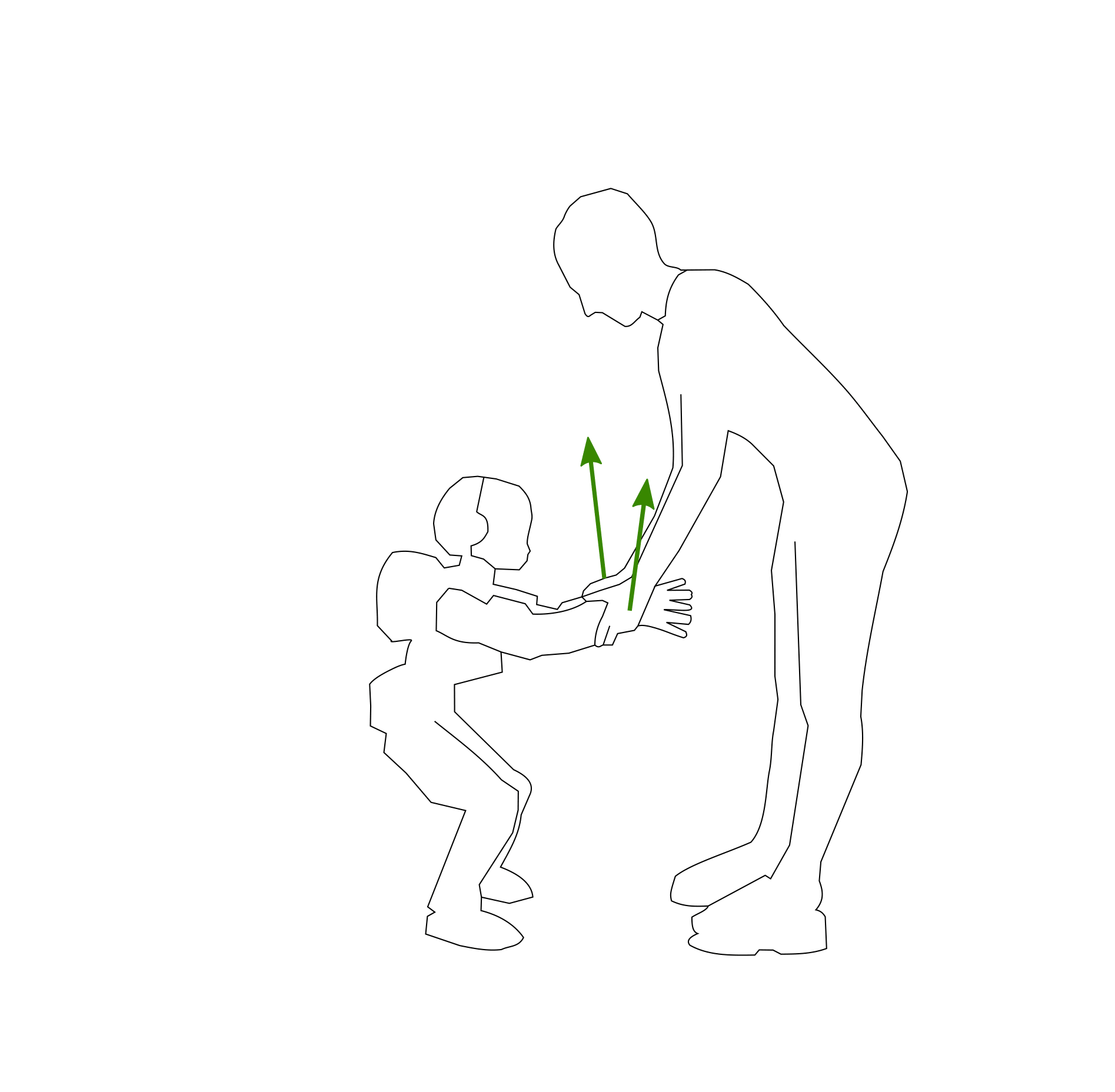}
        \caption{}
        \label{fig:example-icub-help}
    \end{subfigure}
    \caption{HRC example scenarios with a human and a robot involved in physical interactions}
    \label{fig:HRC-scenarios}
\end{figure}

Typical HRC scenarios involve a human and a robotic agent engaged in physical interactions with a common goal of accomplishing a task. Example scenarios of HRC are shown in Fig.~\ref{fig:HRC-scenarios} where a human is helping a manipulator to pick an object or a humanoid robot to stand up by exerting an external wrench. During such scenarios, an intuitive robot behavior is to leverage human assistance and achieve its task quicker. This paper aims at endowing robots with the capabilities to advance along a reference trajectory by leveraging assistance provided during HRC.


 Given the task of tracking a reference trajectory, impedance control \cite{hogan1984impedance} \cite{magrini2015control} is one of the most exploited approaches to achieve stable robot behavior while maintaining contacts. It facilitates a compliant and safe physical interaction between the agents in HRC scenarios. Any interaction wrench from the human is handled safely by changing the robot's actual trajectory i.e. the forces and moments are controlled by acting on position and orientation changes. Novel adaptive control schemes are successfully implemented expanding the applicability of impedance control \cite{gopinathan2017user} \cite{li2013impedance} \cite{gribovskaya2011motion}. The quality of interaction is further augmented through online adaptive admittance controls schemes which consider human intent inside the control loop \cite{ranatunga2016adaptive} \cite{lecours2012variable}. These adaptive control schemes endow robots with compliant characteristics that are safe for physically interacting with them. An obvious outcome of such compliance is the momentary deviation from the reference trajectory to accommodate external interactions but the original trajectory is restored when the interaction stops \cite{geravand2013human}. 
 

The concept of trajectory deformation facilitates the adaptation of robot trajectories to handle external perturbations or possible obstacles that are present in the robot's original trajectory. The authors of \cite{pham2015new} use affine transformations on parts of the motion trajectory which ensures preserving affine-invariant features of the original trajectory like line smoothness and velocity. More recently, the authors of \cite{losey2018trajectory} demonstrated optimal trajectory deformation through constrained optimization of an energy function ensuring the minimum-jerk profile. Although the resulting deformed trajectories are optimal, a main limitation in the above works is that the speed of the task is unchanged.

The framework of dynamic movement primitives (DMP) is a class of dynamical systems that enables task representation by a set of differential equations \cite{ijspeert2013dynamical}. This facilitated numerous works on robot learning by demonstration and successfully achieved motion planning, on-line trajectory modification, imitation learning and skill transfer \cite{7989877} \cite{6386006} \cite{ijspeert2002movement}. More interestingly, the authors of \cite{nemec2013velocity} \cite{nemec2018human} present an intuitive approach to HRC in which the task representation is captured through speed-scaled dynamic motion primitives that allows changing the speed of the task through physical interactions. One of the main limitations in these approaches is the general applicability for trajectory generation without the foremost step in which the desired movement that solves a given task e.g. pick and place has to be learned through direct imitation or kinesthetic guiding of the robot. Furthermore, choosing the right values for all the parameters involved is rather complicated.

Physical interactions during HRC are often intentional and can provide informative insights that can augment the task completion \cite{bajcsy2017learning}. Consider an example case of a robot moving along a given Cartesian reference trajectory performing a pick and place task. An intuitive interaction of a human with the intention to speed up the robot motion is to apply forces in the robot's desired direction. Under such circumstances, traditionally, the robot can either render a compliant behavior through impedance/admittance control or switch to gravity compensation mode that allows the human to move the robot freely (compromising task accuracy). Instead, a more intuitive behavior is to advance further along the reference trajectory and complete the task quicker. This motivates us to propose a trajectory advancement approach through which the robot can advance along the reference trajectory leveraging assistance from physical interactions.


The rest of the paper is organized as follows: Section \ref{background} presents the basic notation, modeling used in this paper followed by a brief description of the classical feedback linearization control approach with the limitations in HRC scenarios and then present the problem statement. Section \ref{method} describes our approach to constructing a parametrized reference trajectory, mathematical definition of helpful interaction and the proposition of trajectory advancement. Description of the experiments conducted and related specifications are laid down in section \ref{experiments} followed by a discussion of the results in section \ref{results} and conclusions in section \ref{conclusions}.
\section{BACKGROUND}
\label{background}

\subsection{Notation \& Modeling}

\begin{itemize}
    \item The inertial frame of reference is denoted by $A$, with $z$-axis pointing against gravity.
    \item The constant $g$ denotes the norm of the gravitational acceleration.
    \item The robotic system is considered to be \textit{floating base} \cite{Featherstone2007} that has $n+1$ rigid body links connected through $n$ joints.
    \item The configuration space of a \textit{free-floating} system is characterized by the \textit{joint positions} and the \textit{floating base frame} $F$. It is defined as a set of elements with $6$ dimensions representing the \textit{floating base} and the total number of joints $n$. Hence, it lies on the Lie group $\mathbb{Q} = \mathbb{R}^3 \times SO(3) \times \mathbb{R}^n$. 
    \item An element in the configuration space is denoted by $q = (q_b,s) \in \mathbb{Q}$, which consists of pose of the \textit{base frame} $q_b = (^{\scalebox{\framesu_size}{A}}p_{\scalebox{\framesu_size}{F}}, ^{\scalebox{\framesu_size}{A}}R_{\scalebox{\framesu_size}{F}}) \in \mathbb{R}^3 \times SO(3)$ where  $^{\scalebox{\framesu_size}{A}}p_{\scalebox{\framesu_size}{F}} \in \mathbb{R}^3$ denotes the position of the base frame with respect to the inertial frame; $^{\scalebox{\framesu_size}{A}}R_{\scalebox{\framesu_size}{F}} \in SO(3)$ denotes the rotation matrix representing the orientation of the base frame with respect to the inertial frame; and the joint positions vector $s \in \mathbb{R}^n$ captures the topology of the robot.
    \item The robot velocity is characterized by the linear and angular velocity of the \textit{base frame} along with the \textit{joint velocities}. The configuration velocity space lies on the Lie group $\mathbb{V} = \mathbb{R}^3 \times \mathbb{R}^3 \times \mathbb{R}^n$ and an element $\nu \in \mathbb{V}$ is defined as $\nu = (\prescript{\mathcal{I}}{}{\mathrm{v}}_\mathcal{B}, \dot{s})$ where $\prescript{\mathcal{I}}{}{\mathrm{v}}_\mathcal{B}=(\prescript{\mathcal{I}}{}{\dot{p}}_\mathcal{B}, \prescript{\mathcal{I}}{}{\omega}_\mathcal{B}) \in \mathbb{R}^6$ denotes the linear and angular velocity of the \textit{base frame} expressed with respect to the inertial frame, and $\dot{s} \in \mathbb{R}^n$ denotes the joint velocities.
\end{itemize}


The equations of motion of a \textit{floating base} robotic system are described by,

\begin{equation}
	M(q) \dot{\nu} + C(q,\nu) \nu + G(q) = B {\tau} + J_c^T f^*
	\label{eq:equations-of-motion}
\end{equation}

where, $M \in \mathbb{R}^{n+6 \times n+6}$ is the mass matrix, $C \in \mathbb{R}^{n+6 \times n+6}$ is the Coriolis matrix, $G \in \mathbb{R}^{n+6}$ is the gravity term, $B = (0_{n \times 6},1_n)^T$ is a selector matrix, ${\tau}  \in \mathbb{R}^{n}$ is a vector representing the robot's joint torques, $f^* \in \mathbb{R}^{6n_c}$ represents the external wrenches acting on $n_c$ contact links of the robot, and $J_c \in \mathbb{R}^{n+6 \times 6n_c}$ is the contact jacobian. 

\subsection{Classical Feedback Linearization Control}

Consider the problem of Cartesian trajectory tracking by a link of the robot where $x_d(t), \dot{x}_d(t), \ddot{x}_d(t) \in \mathbb{R}^6$ denote the desired position, velocity and acceleration in Cartesian space, parametrized in time $t$. Now, $\dot{\widetilde{x}} = \dot{x}(t) - \dot{x}_d(t)$ is the velocity tracking error to be minimized. The robot link's actual velocity has a linear map to the robot's velocity through the Jacobian matrix $J(q) \in \mathbb{R}^{6 \times n+6}$, i.e.

\begin{equation}
	\dot{x}(t) = J (q) \nu
	\label{eq:linear-jacobian-mapping}
\end{equation}

The control objective for the tracking task is defined as,

\begin{equation}
	\label{eq:cartesian-control-objective}
	\ddot{x} = \ddot{x}^* := \ddot{x}_d - K_D \ \dot{\widetilde{x}} - K_P \int_0^t \dot{\widetilde{x}} du, \quad K_D, K_P > 0
\end{equation}
where $K_P, K_D \in \mathbb{R}^{6 \times 6}$ are positive symmetric feedback matrices. According to the classical feedback linearisation approach \cite{khalil2004modeling} we can find the robot joint torques $\tau$ such that the control objective \eqref{eq:cartesian-control-objective} is satisfied and the trajectory tracking error is minimized. The robot control torques necessary for trajectory tracking with the desired dynamics directed by Eq.~\eqref{eq:cartesian-control-objective} are obtained using Eq.~\eqref{eq:linear-jacobian-mapping}. On differentiating $\dot{x}(t)$ we get the following relation,

\begin{equation}
	\ddot{x}(t) = J \dot{\nu} + \dot{J} \nu
	\label{eq:velocity-differentiaion}
\end{equation}

The quantity $\dot{\nu}$ in the above equation is the robot's acceleration that can be derived from the equations of motion \eqref{eq:equations-of-motion} as $\dot{\nu} = M^{-1}[B {\tau} + J_c^T f^* - h]$ where, $h = C(q,\nu) \nu + G(q)$. Using this relation in \eqref{eq:velocity-differentiaion}, we get

\begin{subequations}
    \begin{equation}
	    \ddot{x}(t) = J M^{-1}[B {\tau} + J_c^T f^* - h] + \dot{J} \nu \notag
    \end{equation}
    \begin{equation}
	    \ddot{x}(t) = J M^{-1}B {\tau} + J M^{-1}J_c^T f^* - J M^{-1}h + \dot{J} \nu \notag
    \end{equation}
    \begin{equation}
	     {\tau} = [J M^{-1}B]^{\dagger} [\ddot{x}(t)  - J M^{-1}J_c^T f^* + J M^{-1}h - \dot{J} \nu] \notag
    \end{equation}
\end{subequations}

Now, using the desired dynamics from Eq.~\eqref{eq:cartesian-control-objective}, we compute the control torques as

\begin{equation}
    {\tau} = [J M^{-1}B]^{\dagger} [\ddot{x}^*  - J M^{-1}J_c^T f^* + J M^{-1}h - \dot{J} \nu]
	\label{eq:normal-control-torques}
\end{equation}

On putting in a compact form, we have:

\begin{equation}
    {\tau} = \mathbold{\Delta}^{\dagger} [\ddot{x}^*  - \mathbold{\Omega} f^* + \mathbold{\Lambda}] + N_{\mathbold{\Delta}} \tau_0
	\label{eq:normal-control-torques-compact}
\end{equation}

where 

\begin{itemize}
    \item $ \mathbold{\Delta} = J M^{-1} B \in \mathbb{R}^{6 \times \mathrm{n}} $
    \item $ \mathbold{\Omega} = J M^{-1} J_c^T \in \mathbb{R}^{6 \times 6\mathrm{n}_c}$
    \item $ \mathbold{\Lambda} = J \ M^{-1} h - \dot{J} \nu\in \mathbb{R}^{6}$
    \item $ N_{\mathbold{\Delta}} \in \mathbb{R}^{n \times n}$ is the nullspace projector of $\mathbold{\Delta}$ 
    \item $ \tau_0 \in \mathbb{R}^{n}$ represent torques required to satisfy lower priority tasks in case of redundancy in joint torques
\end{itemize}
    
The above control torques completely cancel out any external wrench applied during physical interactions with the robot. Although this approach is quite robust to external perturbations, it is also limited in facilitating HRC scenarios that require active collaboration between a human partner and a robot \cite{8093992}. 

\subsection{Problem Statement}

Given a reference trajectory to be tracked by a link of the robot, the problem statement can be summarized as how to advance along the reference trajectory by exploiting \textit{helpful} physical interactions with the robot. Accordingly, the main contributions of this work are:

\begin{itemize}
    \item Mathematically defining \textit{helpful} interaction that provides assistance to accomplish a task;
    \item Designing a parametrized reference trajectory and determining the conditions that facilitate advancing along it using the assistance from physical interactions.
\end{itemize}


\section{METHOD}
\label{method}

\subsection{Parametrized Reference Trajectory}

Traditionally, motion control problems involving tracking of a reference trajectory has both spatial dimension, encapsulated in geometric path, and temporal dimension, encapsulated in the dynamic evolution of the geometric path \cite{aguiar2004path}. Accordingly, the reference trajectory is a time ($t$) parametrized curve and the control design drives the system to a specific point in space at a specific pre-defined time. In contrast, the path following problem involves converging to and following a geometric path without any temporal constraints \cite{breivik2005principles}. In this work, we bank on the concepts of path following and design a parametric curve parametrized with a \textit{free parameter} $\psi \in [0, \infty)$. The choice of $\psi$ becomes clear in the subsequent sections. The resulting parametric curve $x_d(\psi)$ is the desired geometric path to be followed spatially by a link of the robot. Assuming that the free parameter is time dependent i.e., $\psi = \psi(t)$, the first and second time derivatives of the path are given as following, 


\begin{subequations}
    \begin{equation}
        \dot{x}_d(\psi, \dot{\psi}) = \partial_\psi x_d(\psi) \ \dot{\psi}
        \label{eq:trajectory-first-derivative}
    \end{equation}
    \begin{equation}
        \ddot{x}_d(\psi, \dot{\psi}, \ddot{\psi}) = \partial_{\psi}^2 x_d(\psi) \ \dot{\psi}^2 + \partial_\psi x_d(\psi) \ \ddot{\psi}
        \label{eq:trajectory-second-derivative}
    \end{equation}
\end{subequations}

\subsection{Interaction Exploitation}

Consider the control objective of trajectory tracking where at each time instant the reference position ($x_d(\psi))$, velocity $(\dot{x}_d(\psi, \dot{\psi}))$, and acceleration $(\ddot{x}_d(\psi, \dot{\psi}, \ddot{\psi}))$ are taken from the reference trajectory parametrized in $\psi$. The term $\mathbold{\Omega} f^*$ in the control torques Eq.~\eqref{eq:normal-control-torques-compact} represents the Cartesian resultant acceleration that results under the influence of external interaction wrench $f^*$. Instead of completely cancelling out the effects of external interaction wrench, it is desirable to exploit any \textit{helpful} components to advance along the desired reference trajectory making an active collaboration possible during HRC. More specifically, let us define the \emph{helpful} interaction by decomposing the external wrenches into \textit{parallel} and \textit{perpendicular} components along the desired velocity as,

\begin{subequations}
    \begin{equation}
        \mathbold{\Omega}f^* = \alpha \ \dot{x}_d^{\parallel} + \beta \ \dot{x}_d^{\perp} \notag
    \end{equation}
    \begin{equation}
        \dot{x}_d^{\parallel} = \frac{\dot{x}_d}{\norm{\dot{x}_d}}, \quad \alpha = \frac{\dot{x}_d^T \mathbold{\Omega} f^*}{\norm{\dot{x}_d}} \notag
    \end{equation}
\end{subequations}
where $\dot{x}_d^{\parallel} \in \mathbb{R}^{6}$ is the unit vector along the direction of the desired velocity, $\alpha \in \mathbb{R}$ is the resultant acceleration component projected along the direction parallel to the direction of desired velocity. An intuitive choice for the component $\alpha$ is in the direction of desired velocity i.e. $\alpha > 0$. Accordingly, we define a \textit{correction wrench}\footnote{The name \textit{correction wrench} is an abuse of notation but has an intuitive meaning in conveying the notion of helpful interaction wrench. Also, the units of wrench $[\si{\newton},\si{\newton\meter}]$ are used.} term given by $\alpha \ \dot{x}_d^{\parallel} \in \mathbb{R}^{6} \ \forall \ \alpha > 0$, that represents the \textit{helpful} interaction mathematically.


\subsection{Trajectory Advancement}

\begin{proposition}
\label{proposiiton-update-law}
The time evolution of the free parameter $\psi$ for trajectory advancement leveraging assistance is given by the following update rule,
\begin{equation}
    \dot{\psi} = min \left\{ \dot{\psi}_{upper}, max \left\{ 1, \frac{\dot{x}(t)^T \ \partial_{\psi} x_d(\psi)}{\norm{\partial_{\psi} x_d(\psi)}^2}  \right\}\right\}
    \label{eq:update-rule}
\end{equation}

\end{proposition}

\begin{proof}
    Given the correction wrench term, the desired dynamics for the trajectory tracking task is updated as,

\begin{equation}
	\label{eq:control-objective-updated}
	\ddot{x} = \ddot{x}^* := \ddot{x}_d - K_D \ \dot{\widetilde{x}} - K_P \int_0^t \dot{\widetilde{x}} du + \alpha \ \dot{x}_d^{\parallel} \ \ \forall \ \alpha > 0
\end{equation}

Using the above choice of the desired dynamics, the robot control torques defined in Eq.~\eqref{eq:normal-control-torques-compact} will only compensate for external wrench that is not helpful. Now, consider the following Lypunov function candidate, 

\begin{equation}
	\mathrm{V} = \frac{1}{2} \norm{\dot{x}(t) - \dot{x}_d(\psi, \dot{\psi})}^{2} + \frac{K_P}{2} \norm{\int_{0}^{t}(\dot{x}(t) - \dot{x}_d(\psi, \dot{\psi})) du}^{2}
	\label{eq:lyapunov-function}
\end{equation}

On differentiating $\mathrm{V}$, we get: 

\begin{subequations}
	\begin{equation}
		\dot{\mathrm{V}} = \ \dot{\widetilde{x}}^T \ddot{\widetilde{x}} + \int_{0}^{t} \dot{\widetilde{x}}^T du \ K_p \dot{\widetilde{x}} \notag		
	\end{equation}
	\begin{equation}
        \dot{\mathrm{V}} = \dot{\widetilde{x}}^T [\ddot{\widetilde{x}} + K_p \int_{0}^{t}\dot{\widetilde{x}} \ du] \notag
	\end{equation}
\end{subequations}

Given the updated desired dynamics in Eq.~\eqref{eq:control-objective-updated} we rearrange it as $\ddot{\widetilde{x}} + K_p \int_{0}^{t}\dot{\widetilde{x}} \ du = - K_D \ \dot{\widetilde{x}} + \alpha \ \dot{x}_d^{\parallel}$ and use it in the derivative of the Lyapunov function to obtain the following relation,

\begin{subequations}
	\begin{equation}
        \dot{\mathrm{V}} = - \dot{\widetilde{x}}^T K_D \ \dot{\widetilde{x}} + \dot{\widetilde{x}}^T \alpha \ \dot{x}_d^{\parallel}\  \notag
	\end{equation}
\end{subequations}

According to Lyapunov theory, the stability of the system is ensured when $\dot{\mathrm{V}} \le 0$. Given that $K_D$ is a positive symmetric matrix, the term $- \dot{\widetilde{x}}^T \ K_D \ \dot{\widetilde{x}} \le 0$. So, to ensure the stability of the system i.e. $\dot{\mathrm{V}} \le 0$, the following condition has to be satisfied,

\begin{equation}
    \dot{\widetilde{x}}^T \alpha \ \dot{x}_d^{\parallel} \le 0 \notag
\end{equation}

Considering that $\alpha > 0$ and $\norm{\dot{x}_d} > 0$, the above inequality is equivalent to $\dot{\widetilde{x}}^T \dot{x}_d(\psi, \dot{\psi}) \le 0$,

\begin{subequations}
    \begin{equation}
        (\dot{x}(t) - \dot{x}_d(\psi, \dot{\psi}))^T \dot{x}_d(\psi, \dot{\psi}) \le 0 \notag
    \end{equation}
    \begin{equation}
        \dot{x}(t)^T \dot{x}_d(\psi, \dot{\psi}) - \norm{\dot{x}_d(\psi, \dot{\psi})}^2 \le 0 \notag
    \end{equation}
    \begin{equation}
        \dot{x}(t)^T \partial_{\psi} x_d(\psi)  \dot{\psi} - \norm{\partial_{\psi} x_d(\psi)}^2 \dot{\psi}^2 \le 0 \notag
    \end{equation}
\end{subequations}

Assuming the lower bound $\dot{\psi} \ge 1$, we obtain

\begin{equation}
    \dot{\psi} \ge \frac{\dot{x}(t)^T \partial_{\psi} x_d(\psi)}{\norm{\partial_{\psi} x_d(\psi)}^2}
    \label{eq:sdot-condition}
\end{equation}

The condition in Eq.~\eqref{eq:sdot-condition} reflects the time evolution of the free parameter $\psi$ which helps in advancing along the desired reference trajectory exploiting the external interaction wrenches with the robot. The lower bound value $1$ signifies that the new parametrization is exactly equal to the time parametrized trajectory i.e. $\psi = t$ until any external wrench $f^*$ is applied such that it will help the robot's task. Under the influence of \textit{helpful} external wrench, the value of $\dot{\psi}$ becomes greater than $1$. On integrating/differentiating $\dot{\psi}$ we determine the advancement along the desired reference trajectory,
\begin{equation}
	\psi^* = \int_{t_1}^{t_2} \dot{\psi} \ du, \quad \ddot{\psi}^* = \frac{d\dot{\psi}}{dt} \nonumber
\end{equation}

Now, the updated references for trajectory tracking becomes $x_d(\psi^*), \dot{x}_d(\psi^*,\dot{\psi}), \ddot{x}_d(\psi^*,\dot{\psi},\ddot{\psi}^*)$. Besides, an upper limit $\dot{\psi}_{upper}$ is set to bound the length of advancement along the reference trajectory ensuring safe physical interactions. 



\textbf{Remark:} Strictly speaking, the choice of $\dot{\psi}$ as stated in Eq.~\eqref{eq:update-rule} induces an algebraic loop when applied with the control law Eq.~\eqref{eq:control-objective-updated}. In fact, the updated reference acceleration $\ddot{x}_d(\psi^*,\dot{\psi},\ddot{\psi}^*)$ does depend on the Cartesian acceleration $\ddot{x}$, and, consequently, on the joint torques $\tau$. For this reason, no formal stability statement was claimed in Proposition~\ref{proposiiton-update-law}. From the theoretical point of view, the algebraic loop can be avoided by designing an update rule for $\ddot{\psi}$ rather than $\dot{\psi}$, and by modifying the control law \eqref{eq:control-objective-updated} so that the reference Cartesian acceleration is not compensated anymore. This choice, however, would imply the calculation of $\psi^*$ through double numerical integration of  $\ddot{\psi}$, which may lead to fast divergence of the reference trajectory due to numerical drifts. For this reason, the proposed control solution \eqref{eq:update-rule}-\eqref{eq:control-objective-updated}, despite not being fully theoretically sound, resulted to be more robust when applied in practice. Furthermore, the algebraic loop can be resolved at the implementation level by computing the numerical derivative $\ddot{\psi}^* = \frac{d\dot{\psi}}{dt}$ with one time step of delay, and/or by low-pass filtering the signal to also attenuate the effect of numerical noise. Driven by these motivations we used Eq.~\eqref{eq:update-rule}-\eqref{eq:control-objective-updated} for controlling the robot and verified the closed-loop system stability numerically. The derivation of a controller with proven stability properties is an on going activity that will be carried on in future works.

\end{proof}

\section{EXPERIMENTS}
\label{experiments}

\subsection{Experimental Setup}

\begin{figure}[b]
    \centering
    \begin{subfigure}{0.24\textwidth}
        \centering
        \includegraphics[clip, trim=0 1cm 0 0cm, scale=0.14]{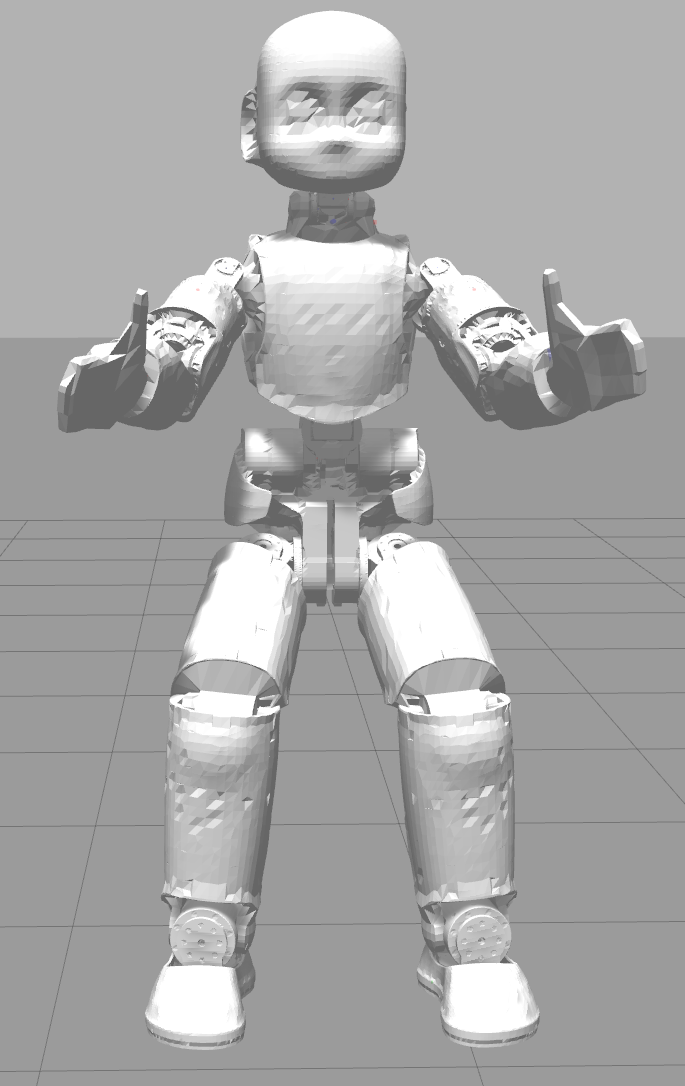}
        \caption{}
        \label{fig:icub-on-pole}
    \end{subfigure}%
    \begin{subfigure}{0.24\textwidth}
        \centering
        \includegraphics[clip, trim=0 1cm 0 0cm, scale=0.132]{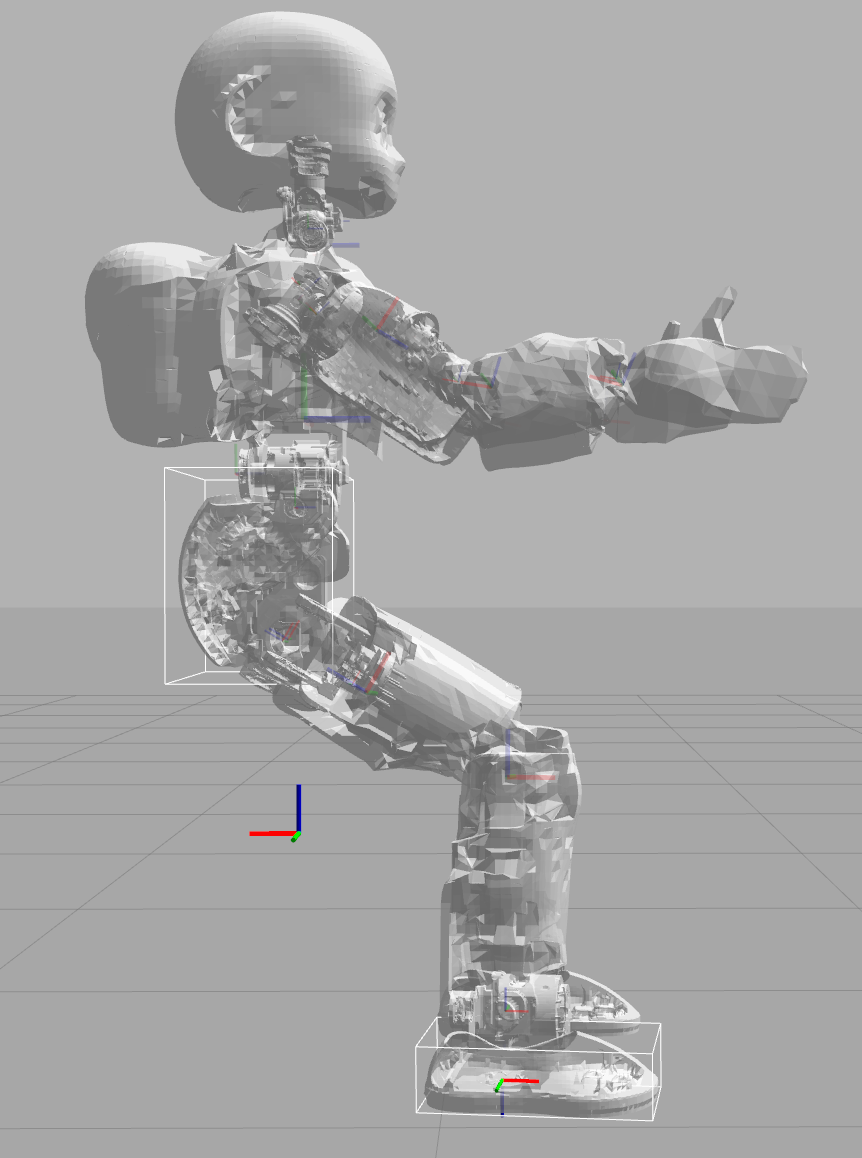}
        \caption{}
        \label{fig:icub-foot-frames}
    \end{subfigure}
    \caption{iCub humanoid robot in gazebo simulation}
    \label{fig:icub}
\end{figure}

The robotic platform considered in our experiments is the iCub humanoid robot \cite{Nataleeaaq1026}. The control objective is to move the \textit{ right foot} of the robot along the desired reference trajectory. The leg of the robot has $3$ joints at the hip, $1$ joint at the knee, and $2$ joints at the ankle. For the sake of intuition and page limitation, this work considers only one dimensional (1D) trajectory in $x$ direction. The reference trajectory is a sinusoidal function of amplitude $0.05 \si{\meter}$ with frequency $0.1 \si{\hertz}$ and is designed to have minimum jerk profile \cite{kyriakopoulos1988minimum}. Concerning the task of trajectory tracking with the leg, the robot base is fixed on a pole as shown in Fig.~\ref{fig:icub-on-pole}. The link frame associated with the right foot of the robot and the inertial frame of reference (shown under the pelvis of the robot) are highlighted in Fig.~\ref{fig:icub-foot-frames}.

\begin{figure}[t]
    \centering
    \begin{subfigure}{0.49\textwidth}
        \centering
        \includegraphics[clip, trim=0.75cm 2.75cm 4.75cm 2.75cm, scale=0.1225]{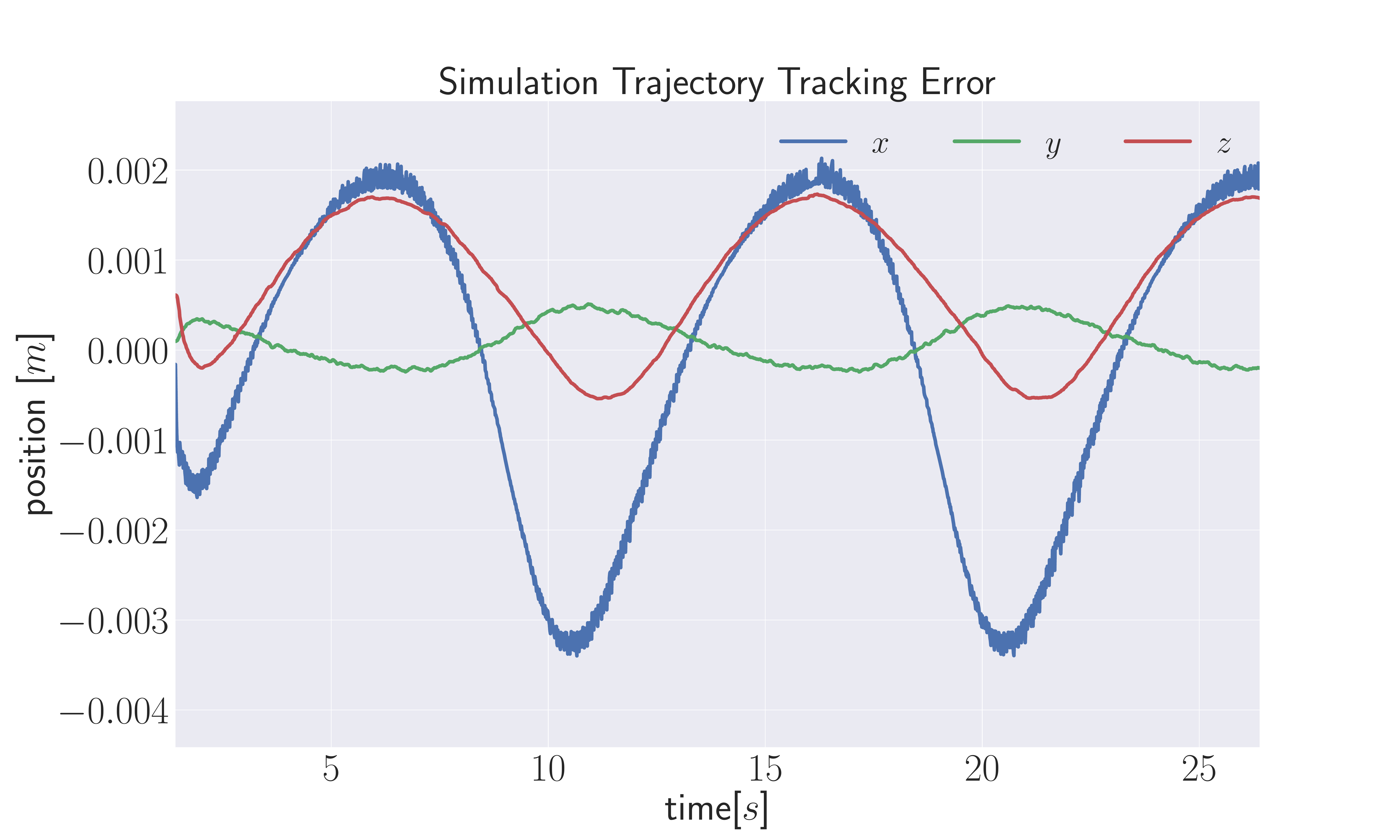}
    \end{subfigure}
    \begin{subfigure}{0.49\textwidth}
        \centering 
        \includegraphics[clip, trim=0.75cm 0.5cm 4.75cm 2.75cm, scale=0.1225]{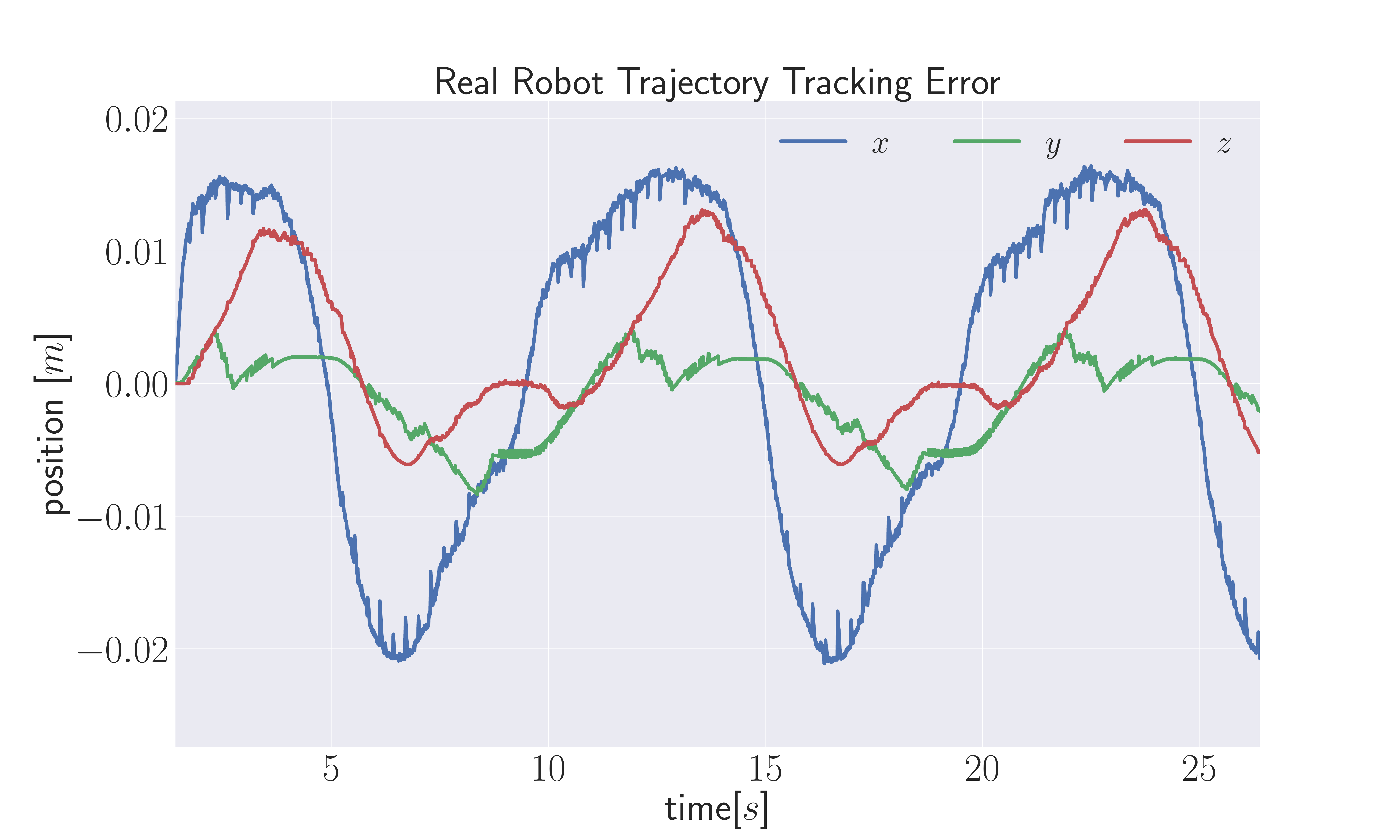}
    \end{subfigure}
    \caption{Trajectory tracking error under no external wrenches}
    \label{fig:normal-trajectory-tracking-1d-x}
\end{figure}

Experiments are carried out in both Gazebo simulation and on the real robot. The controller is implemented in Matlab Simulink, using whole-body toolbox \cite{RomanoWBI17Journal}, as a stack-of-tasks controller with trajectory tracking as the primary objective. The controller gains are tuned to achieve good trajectory tracking both in simulation and on the real robot as highlighted in Fig.~\ref{fig:normal-trajectory-tracking-1d-x}. The trajectory tracking error in the case of simulation is very small and can be attributed to numerical instability of the dynamics integration in Gazebo simulation and numerical noise in measurements. On the other hand, the trajectory tracking error on the real robot is certainly higher than in simulations owing to several unmodeled effects such as joint friction which are prominent on the real robot. Friction induces phase delays in following the desired trajectory resulting in higher tracking error. The upper limit $\dot{\psi}_{upper}$ is set to $10$ for experiments both in simulation and on the real robot.

\subsection{Wrench Classification}

The iCub robot has a force-torque sensor embedded at the end-effector considered for the experiments i.e. the right foot. Instead of reading the sensor measurements directly in sensor frame, the wrench measurements are expressed with a frame that has the origin of the end-effector frame and the orientation of the inertial frame of reference \cite{nori2015icub}. An external wrench applied to the link of the robot is classified, in this work, in two ways: 

\begin{itemize}
    \item \textit{Assistive wrench} if the external wrench has a vector component along the desired direction of motion
    \item \textit{Agnostic wrench} if the external wrench does \textit{not} have vector components along the desired direction of motion
\end{itemize}
 
\begin{figure}[t]
    \centering
    \begin{subfigure}{0.167\textwidth}
        \centering
        \includegraphics[clip, trim=0 1.5cm 0 0.5cm, scale=0.09]{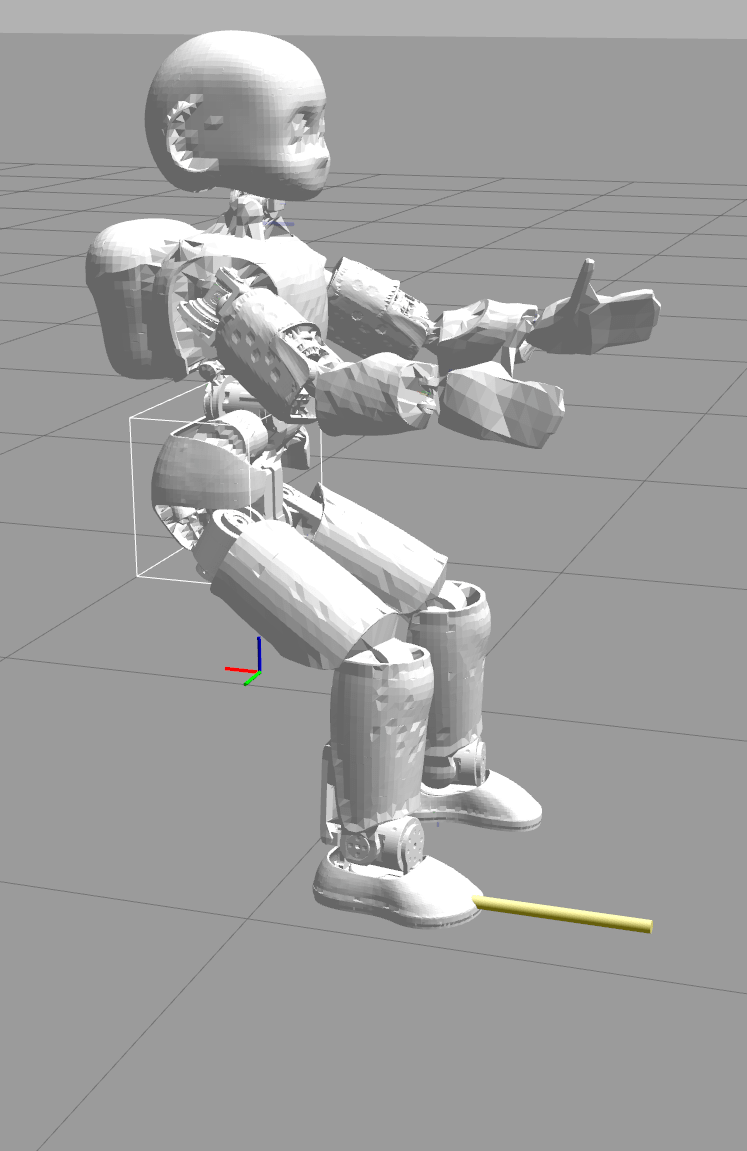}
        \vspace*{-0.2cm}
        \caption{}
        \label{fig:assistive-wrench-1}
    \end{subfigure}%
    \begin{subfigure}{0.167\textwidth}
        \centering
        \includegraphics[clip, trim=0 1.5cm 0 0.5cm, scale=0.09]{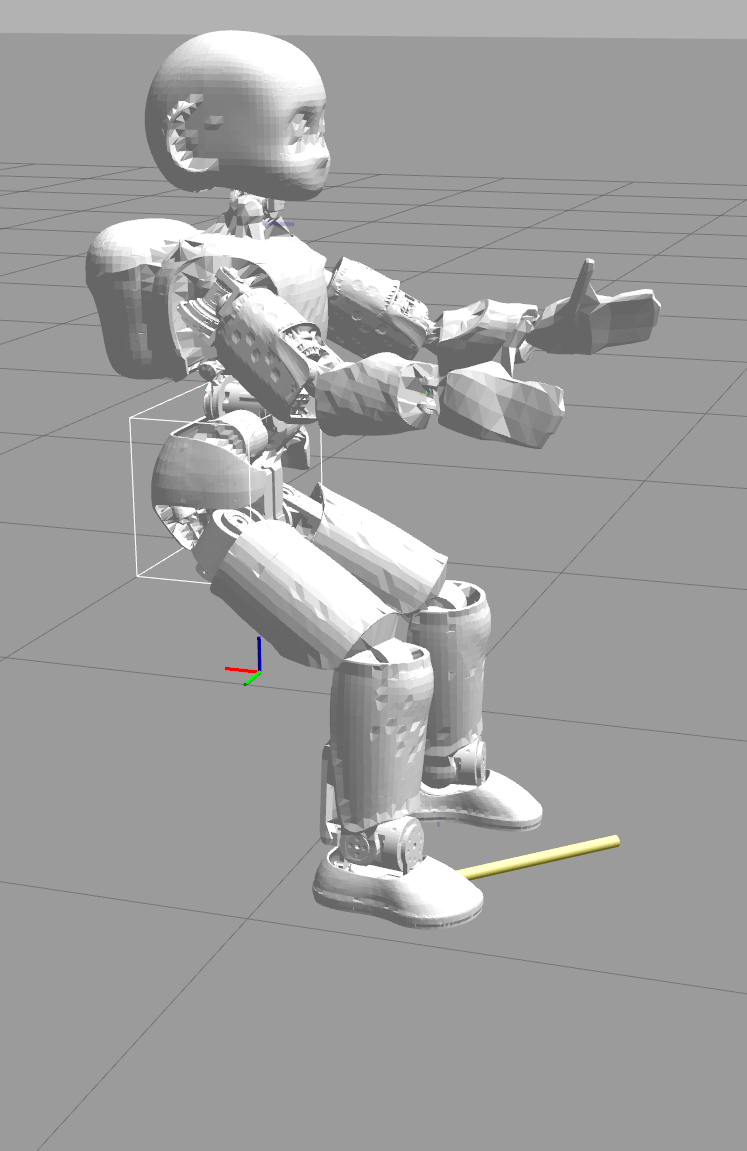}
        \vspace*{-0.2cm}
        \caption{}
        \label{fig:assistive-wrench-2}
    \end{subfigure}%
    \begin{subfigure}{0.167\textwidth}
        \centering
        \includegraphics[clip, trim=0 1.5cm 0 0.5cm, scale=0.09]{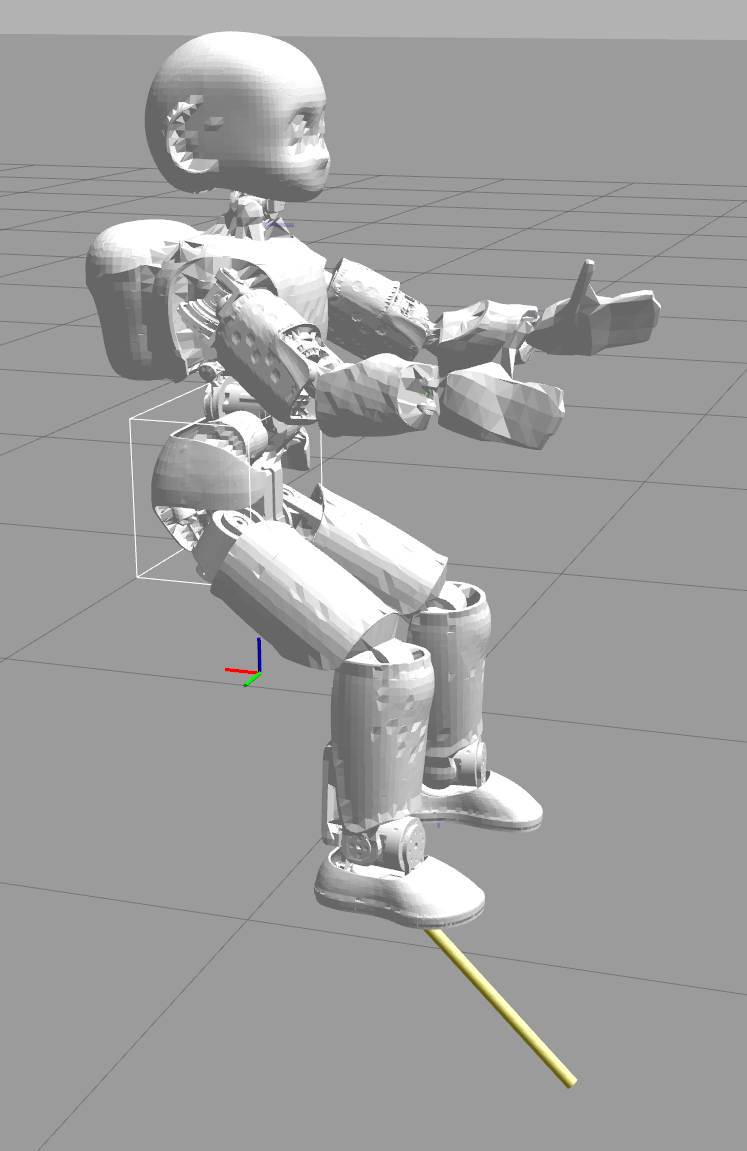}
        \vspace*{-0.2cm}
        \caption{}
        \label{fig:assistive-wrench-3}
    \end{subfigure}
    \begin{subfigure}{0.167\textwidth}
        \centering
        \includegraphics[clip, trim=0 0.5cm 0 0.5cm, scale=0.09]{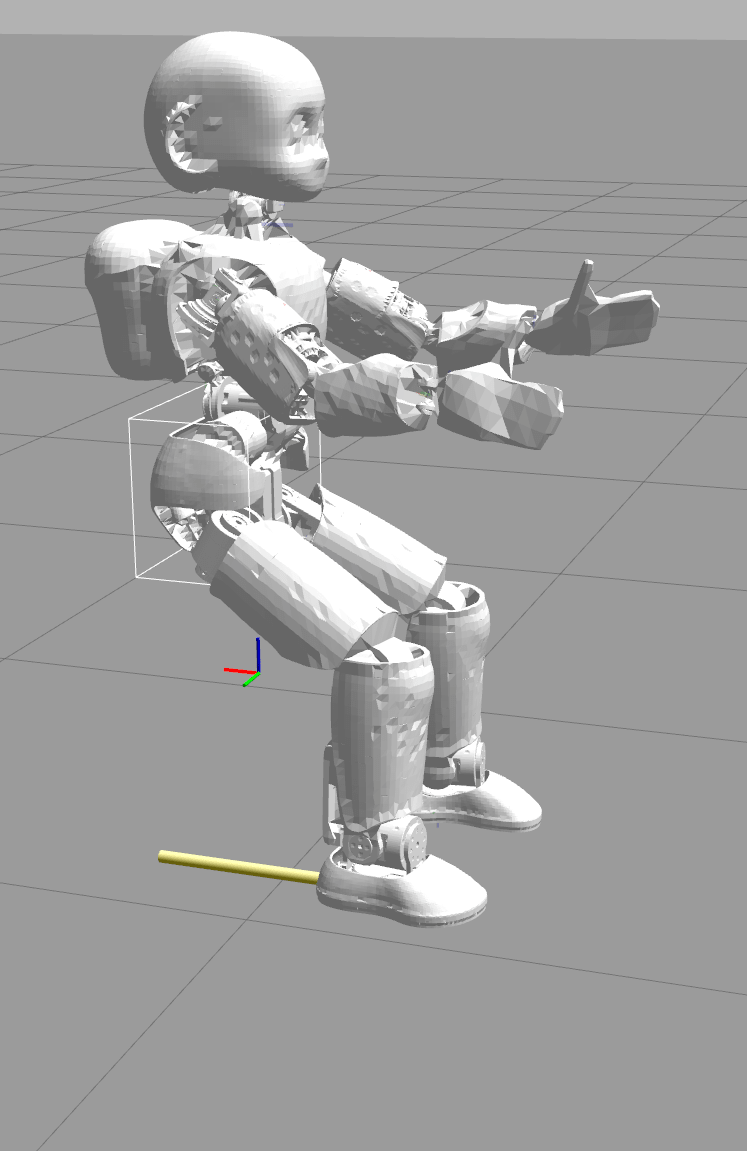}
        \vspace*{-0.2cm}
        \caption{}
        \label{fig:agnostic-wrench-1}
    \end{subfigure}%
    \begin{subfigure}{0.167\textwidth}
        \centering
        \includegraphics[clip, trim=0 0.5cm 0 0.5cm, scale=0.09]{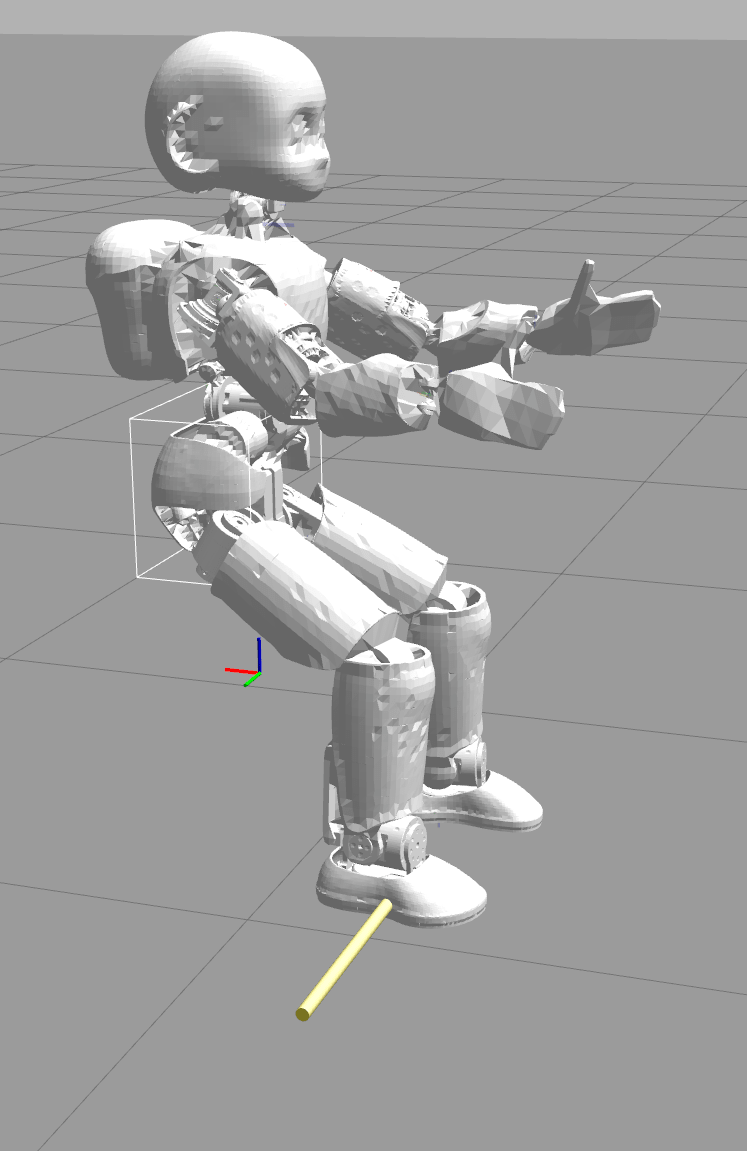}
        \vspace*{-0.2cm}
        \caption{}
        \label{fig:agnostic-wrench-2}
    \end{subfigure}%
    \begin{subfigure}{0.167\textwidth}
        \centering
        \includegraphics[clip, trim=0 0.5cm 0 0.5cm,scale=0.09]{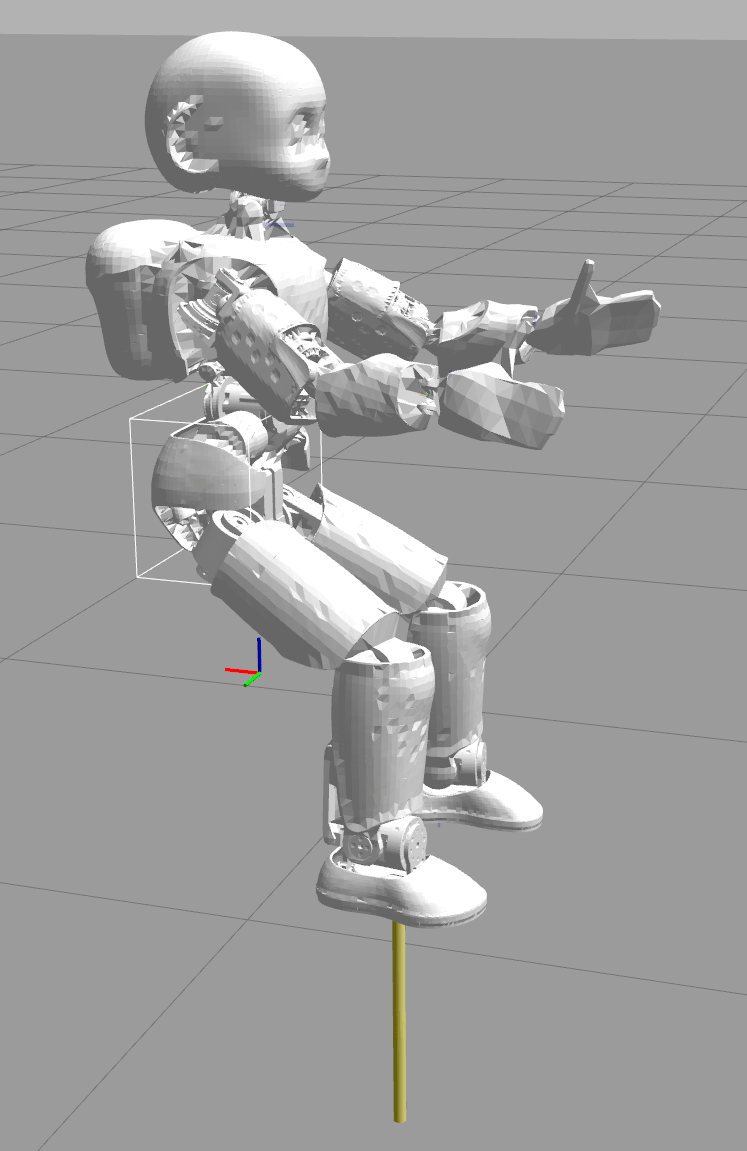}
        \vspace*{-0.2cm}
        \caption{}
        \label{fig:agnostic-wrench-3}
    \end{subfigure}
    \vspace*{-0.1cm}
    \caption{External interaction wrench classification examples when desired direction of motion in positive $x$-direction}
    \label{fig:wrench-classification}
\end{figure}

Examples of external wrench classification are highlighted in Fig.~\ref{fig:wrench-classification}. Considering that the desired direction of motion for the foot is in \textit{positive} $x$-direction with respect to the inertial frame of reference, the external wrenches shown in Fig.~\ref{fig:assistive-wrench-1}~\ref{fig:assistive-wrench-2}~\ref{fig:assistive-wrench-3} are assistive wrenches as they have a vector component along the positive $x$-direction. Similarly, the external wrenches shown in Fig.~\ref{fig:agnostic-wrench-1}~\ref{fig:agnostic-wrench-2}~\ref{fig:agnostic-wrench-3} are agnostic wrenches as they do \textit{not} have any vector component along the positive $x$-direction.
 
Concerning the experiments conducted in Gazebo simulation environment, wrench is applied through a plugin \cite{hoffman2014yarp}. Due to the limited operational space of the robot we chose a fixed duration of $0.75 \si{\second}$ for the wrench. Furthermore, the wrench applied has a smooth profile rather than an impulse profile. This is an experimental design choice made to mimic the intentional interaction wrench applied by a human on the real robot during HRC scenarios.
\section{RESULTS}
\label{results}

\subsection{Simulation}

The first set of experiments are performed in Gazebo simulation environment. A set of test wrenches listed in the table \ref{table:simulation-test-wrenches} are applied when the desired direction of motion is along the positive $x$-direction. These test wrench vectors are similar in \textit{direction} to the wrench vectors highlighted in the wrench classification examples Fig.~\ref{fig:wrench-classification}. The first three wrench vectors (a)(b)(c) are classified as assistive wrenches as they have a vector component (highlighted in \textit{blue}) along the desired direction of motion i.e. positive $x$-direction. The next three wrench vectors (d)(e)(f) are classified as agnostic wrenches as they do not have any vector component along the desired direction of motion.

\begin{table}[!ht]
    \centering
    \begin{tabular}{|p{0.1cm}|p{0.2cm}|p{0.2cm}|p{0.2cm}|p{0.2cm}|p{0.2cm}|p{0.2cm}|}
        \cline{2-7}
        \rowcolor{CustomGray} \multicolumn{1}{c|}{\cellcolor{white}} & $f_x$ & $f_y$  & $f_z$  & $\tau_x$  & $\tau_y$  & $\tau_z$ \\
        \hline
        \rowcolor{GreenYellow} (a) & \colorbox{Periwinkle}{$10$} & $0$  & $0$  & $0$  & $0$  & $0$ \\
        \hline
        \rowcolor{GreenYellow} (b) & \colorbox{Periwinkle}{$5$} & $10$  & $0$  & $0$  & $0$  & $0$ \\
        \hline
        \rowcolor{GreenYellow} (c) & \colorbox{Periwinkle}{$5$} & $0$  & $10$  & $0$  & $0$  & $0$ \\
        \hline
        \rowcolor{Apricot} (d) & $-10$ & $0$  & $0$  & $0$  & $0$  & $0$ \\
        \hline
        \rowcolor{Apricot} (e) & $0$ & $-10$  & $0$  & $0$  & $0$  & $0$ \\
        \hline
        \rowcolor{Apricot} (f) & $0$ & $0$  & $10$  & $0$  & $0$  & $0$ \\
        \hline
    \end{tabular}
    \caption{Test wrenches applied in gazebo simulation}
    \label{table:simulation-test-wrenches}
\end{table}

\begin{figure*}[t]
    \centering
    \begin{subfigure}{0.5\textwidth}
        \centering
        \includegraphics[clip, trim=1cm 0.5cm 4.5cm 1.5cm, scale=0.125]{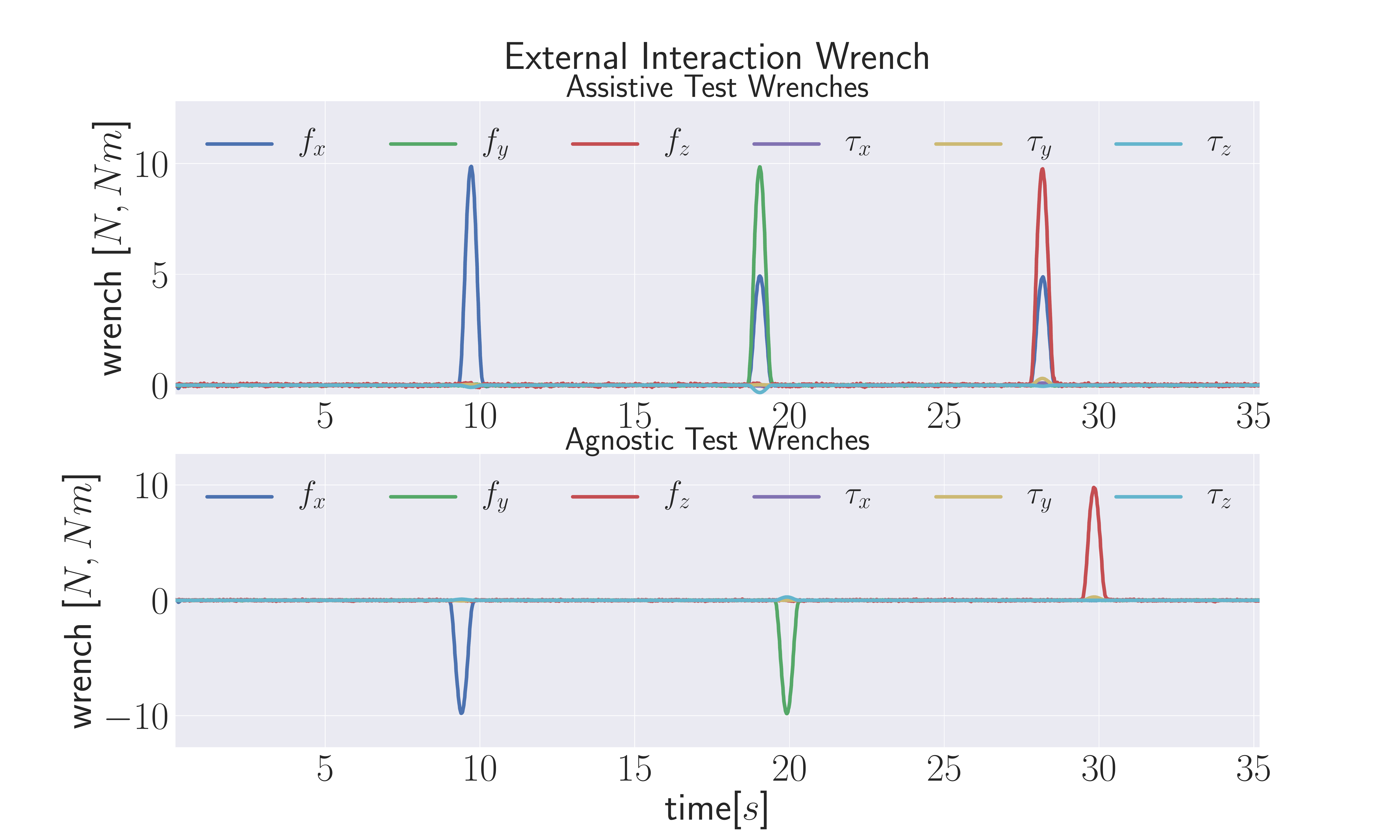}
        \caption{\hspace*{-7.5mm}}
        \label{fig:simulation-external-interaction-wrench-x}
    \end{subfigure}%
    \begin{subfigure}{0.5\textwidth}
        \centering
        \includegraphics[clip, trim=1cm 0.5cm 4.5cm 1.5cm, scale=0.125]{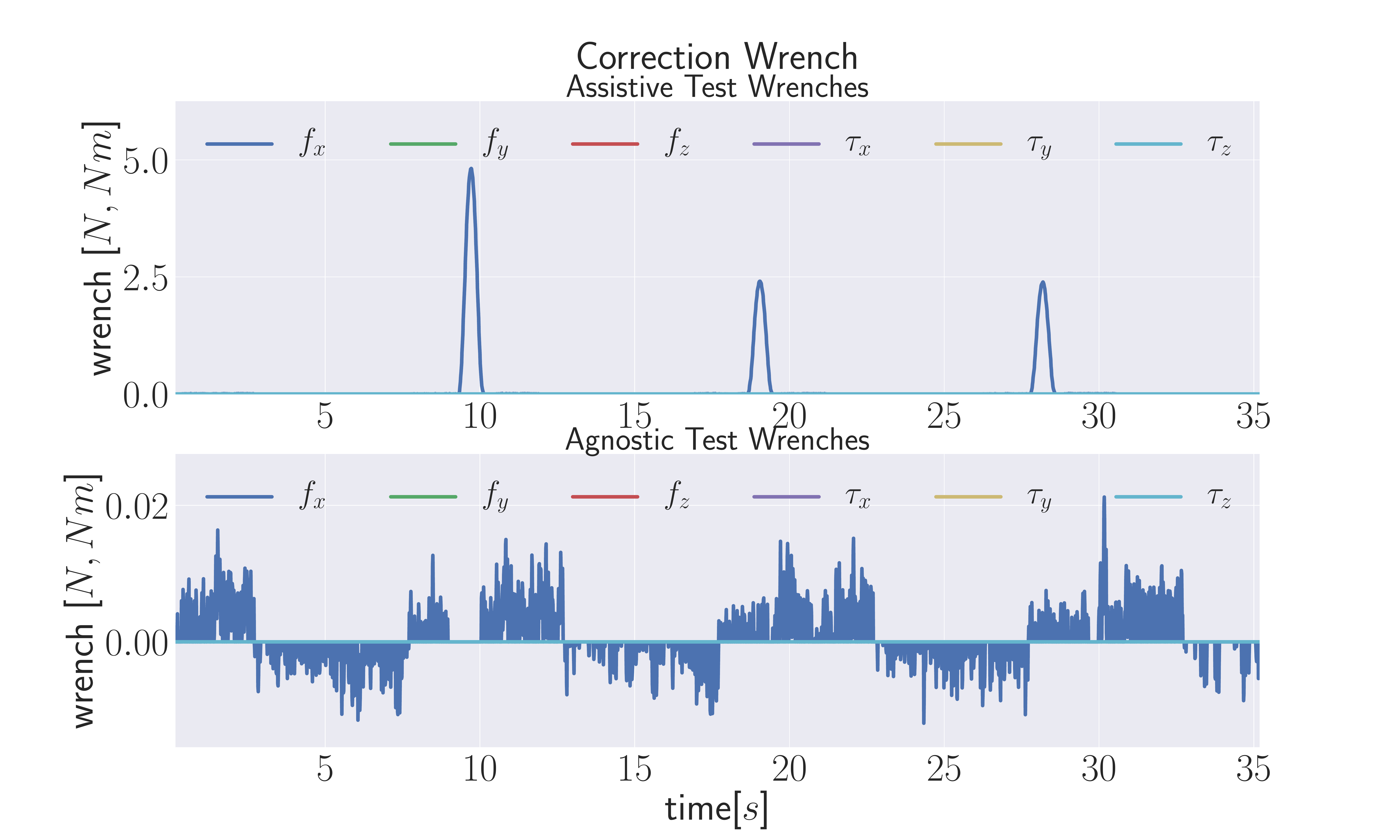}
        \caption{\hspace*{-7.5mm}}
        \label{fig:simulation-correction-wrench-x}
    \end{subfigure}
    \begin{subfigure}{0.5\textwidth}
        \centering
        \includegraphics[clip, trim=1cm 0.5cm 4.5cm 2.5cm, scale=0.125]{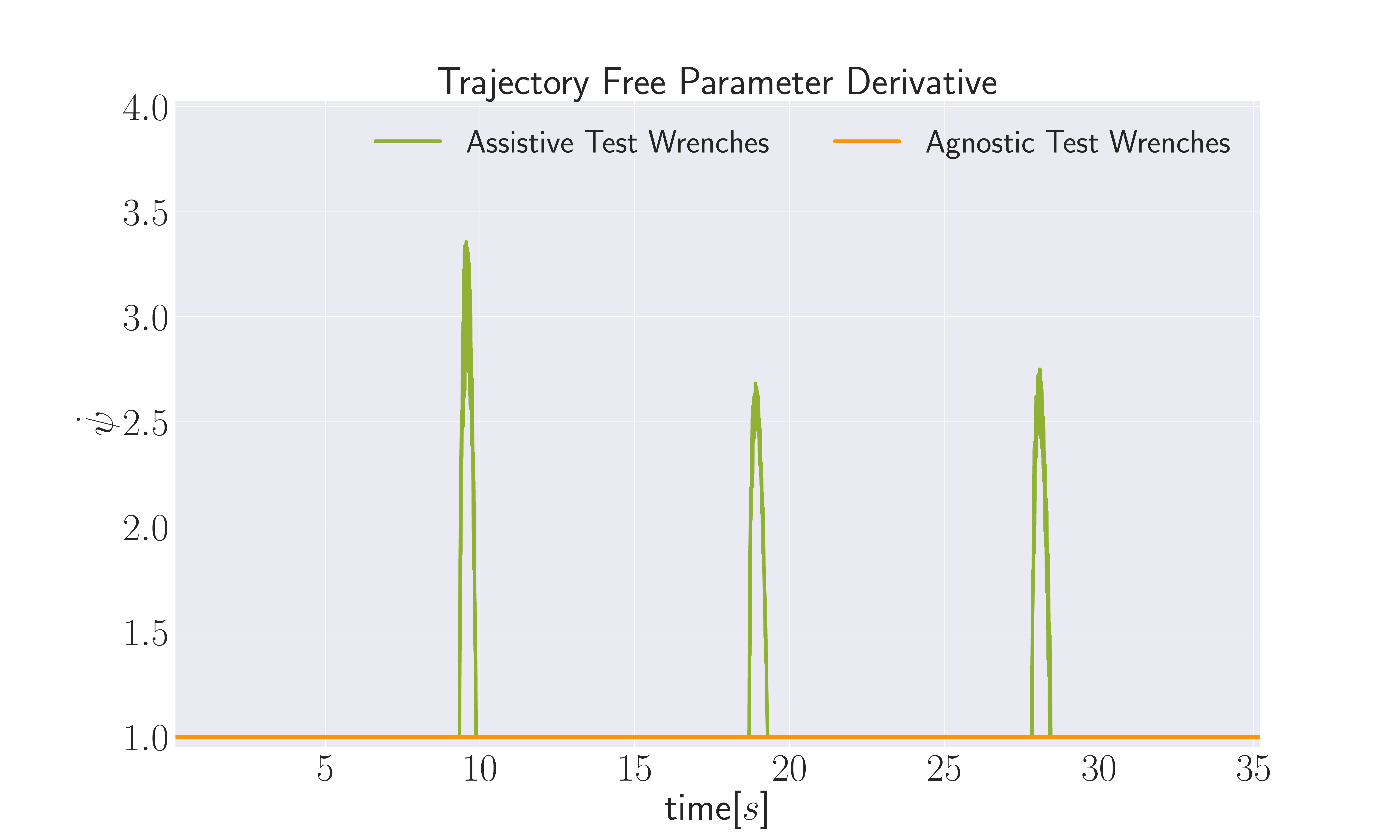}
        \caption{\hspace*{-7.5mm}}
        \label{fig:simulation-sdotvalue-x}
    \end{subfigure}%
    \begin{subfigure}{0.5\textwidth}
        \centering
        \includegraphics[clip, trim=1cm 0.5cm 4.5cm 2.5cm, scale=0.125]{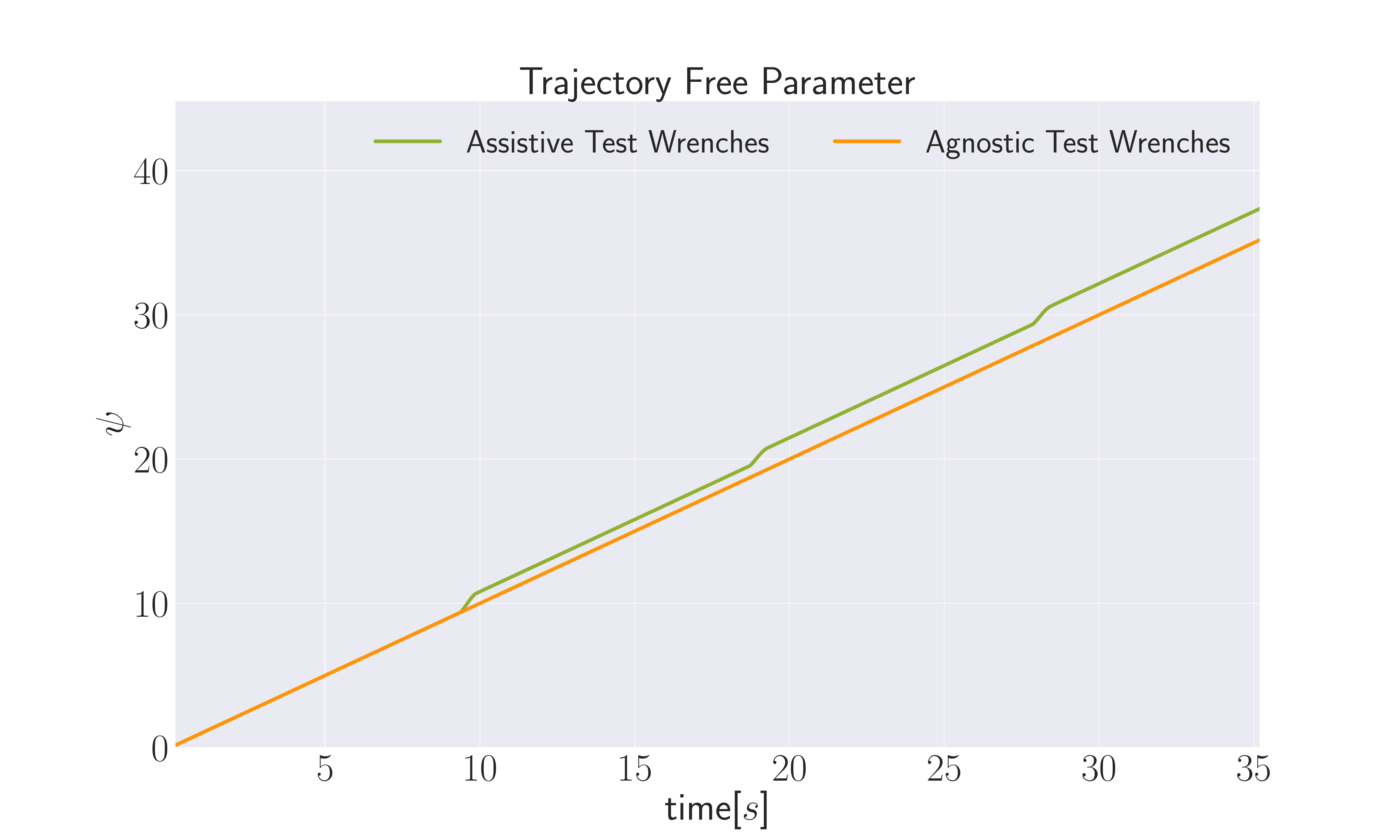}
        \caption{\hspace*{-7.5mm}}
        \label{fig:simulation-svalue-x}
    \end{subfigure}
    \begin{subfigure}{0.5\textwidth}
        \centering
        \includegraphics[clip, trim=1cm 0.5cm 4.5cm 2.5cm, scale=0.125]{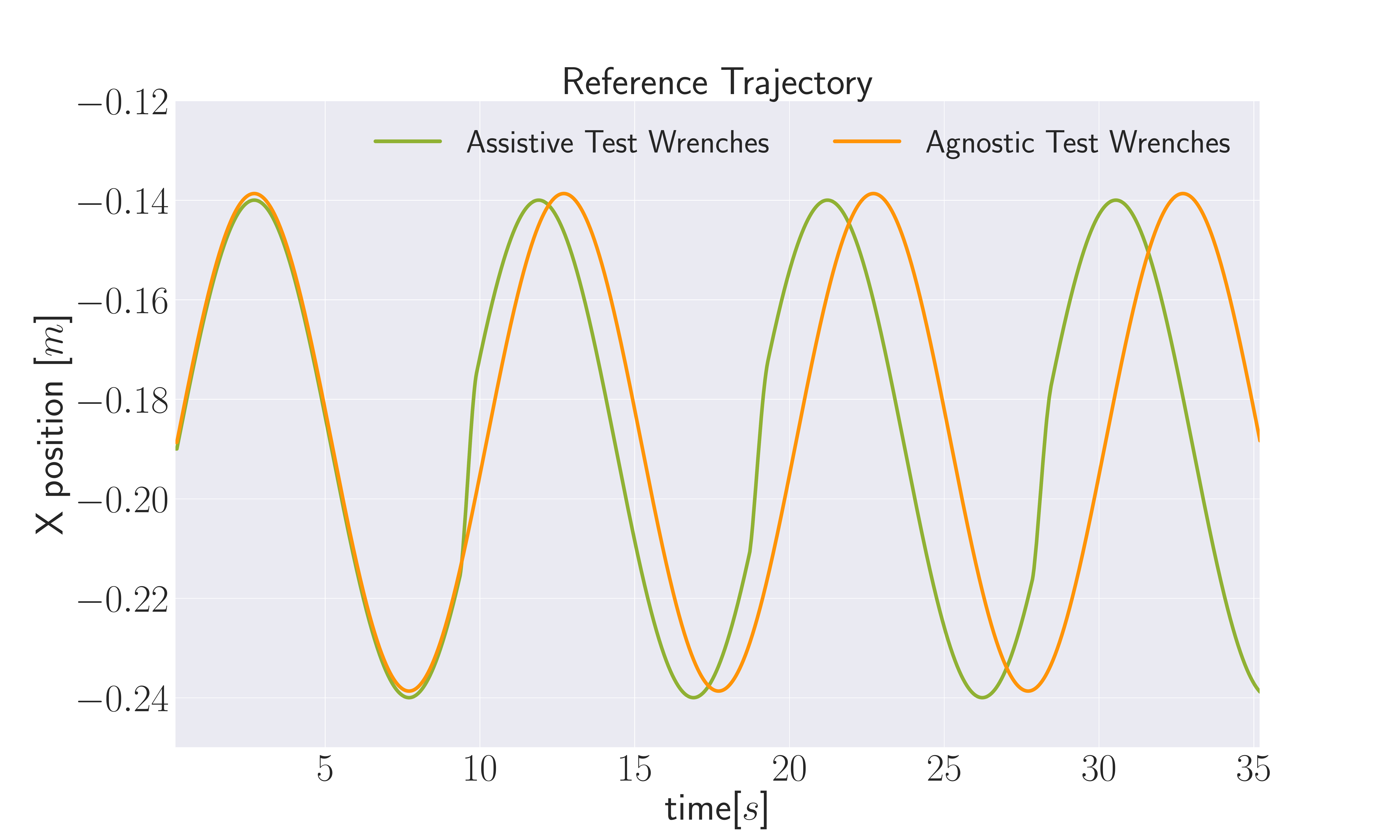}
        \caption{\hspace*{-7.5mm}}
        \label{fig:simulation-reference-trajectory-x}
    \end{subfigure}%
    \begin{subfigure}{0.5\textwidth}
        \centering
        \includegraphics[clip, trim=1cm 0.5cm 4.5cm 2.5cm, scale=0.125]{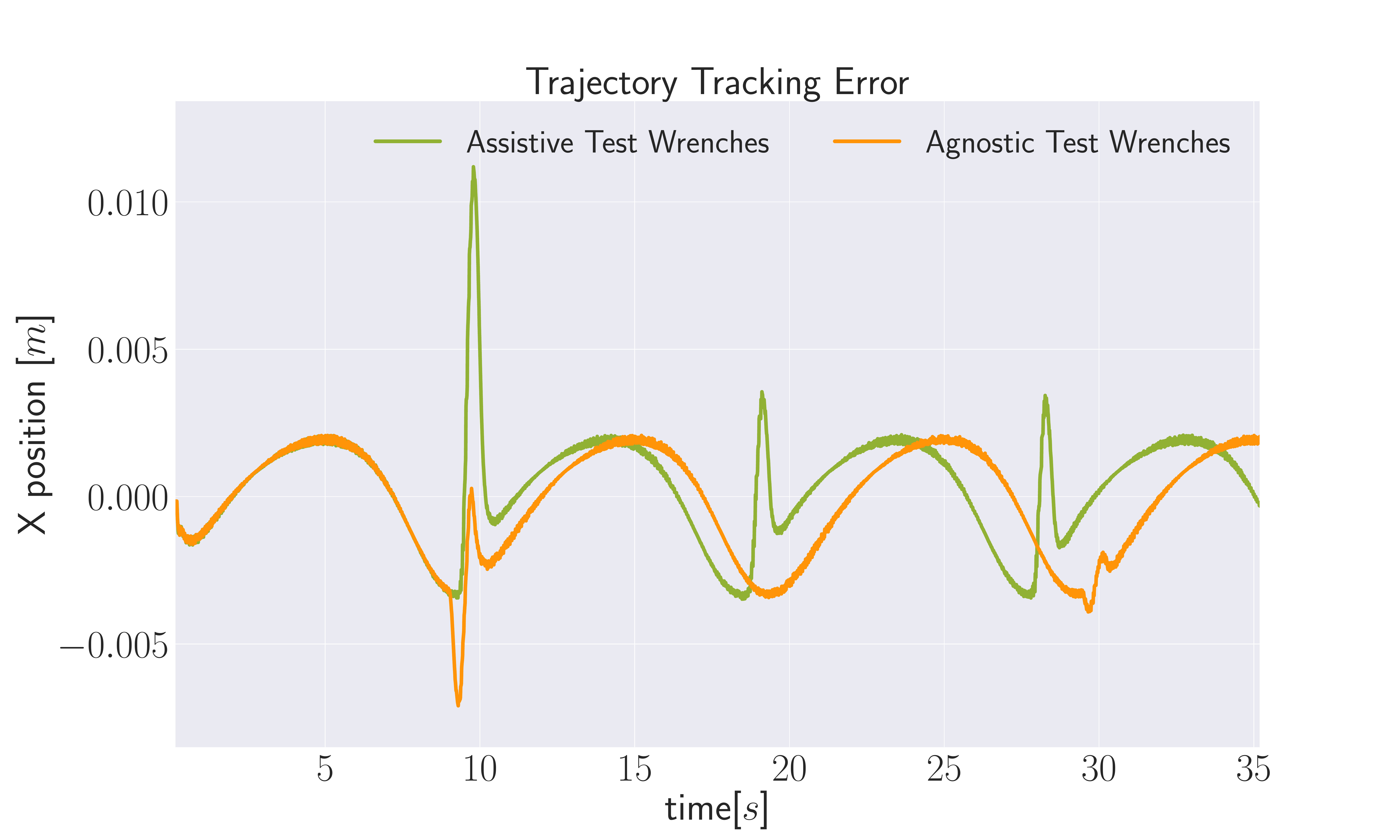}
        \caption{\hspace*{-7.5mm}}
        \label{fig:simulation-trajectory-tracking-error-x}
    \end{subfigure}
    \caption{Gazebo simulation 1D trajectory advancement along $X$-direction}
    \label{fig:simulation-velocity-modulation-1d-x}
\end{figure*}

The results of experiments in Gazebo simulation under the application of the test wrenches listed in table \ref{table:simulation-test-wrenches} are highlighted in Fig.~\ref{fig:simulation-velocity-modulation-1d-x}. The external interaction wrench experienced by the right foot link of the robot are shown in Fig.~\ref{fig:simulation-external-interaction-wrench-x} and the \textit{correction wrench} that is considered towards trajectory advancement is shown in Fig.~\ref{fig:simulation-correction-wrench-x}. In the case of agnostic wrenches, the correction wrench terms are insignificant and they are present due to the noise in the wrench estimation \cite{nori2015icub}. The reference trajectory is similar to a time parametrized trajectory i.e., $\psi = t$ until any helpful wrench is applied to the end-effector. Under the influence of assistive wrenches, the derivative of the trajectory free parameter $\dot{\psi}$ changes as highlighted in Fig.~\ref{fig:simulation-sdotvalue-x} and the corresponding trajectory advancement is reflected as an increase in $\psi$ as seen in Fig.~\ref{fig:simulation-svalue-x}. Accordingly, the reference is advanced further along the reference trajectory as shown in Fig.~\ref{fig:simulation-reference-trajectory-x}.

The trajectory tracking error is slightly more when the reference trajectory is updated under the influence of the assistive wrenches however the error magnitude is of low order as highlighted in Fig.~\ref{fig:simulation-trajectory-tracking-error-x} proving that the task of trajectory tracking is achieved reliably by the controller. Another important observation is that the magnitude of change in $\dot{\psi}$ is related to the magnitude of the interaction wrench. The length of advancement under the influence of assistive wrench vector (a) is more than under the influence of assistive wrench vector (b) or (c) from table \ref{table:simulation-test-wrenches}.

\subsection{Real Robot}

\begin{figure*}[t]
    \centering
    \begin{subfigure}{0.5\textwidth}
        \centering
        \includegraphics[clip, trim=1cm 0.5cm 4.5cm 3.5cm, scale=0.125]{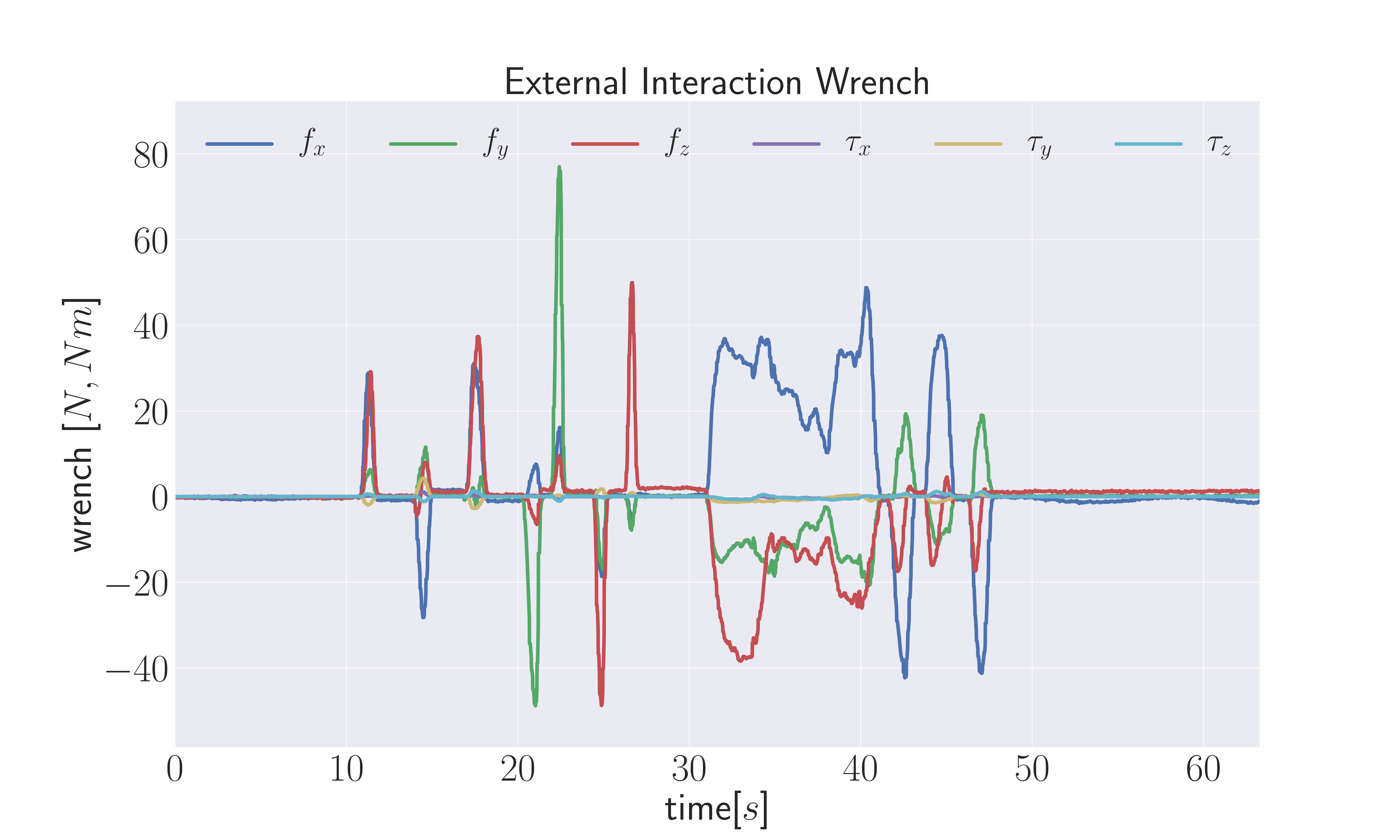}
        \caption{\hspace*{-7.5mm}}
        \label{fig:real-robot-external-interaction-wrench-x}
    \end{subfigure}%
    \begin{subfigure}{0.5\textwidth}
        \centering
        \includegraphics[clip, trim=1cm 0.5cm 4.5cm 3.5cm, scale=0.125]{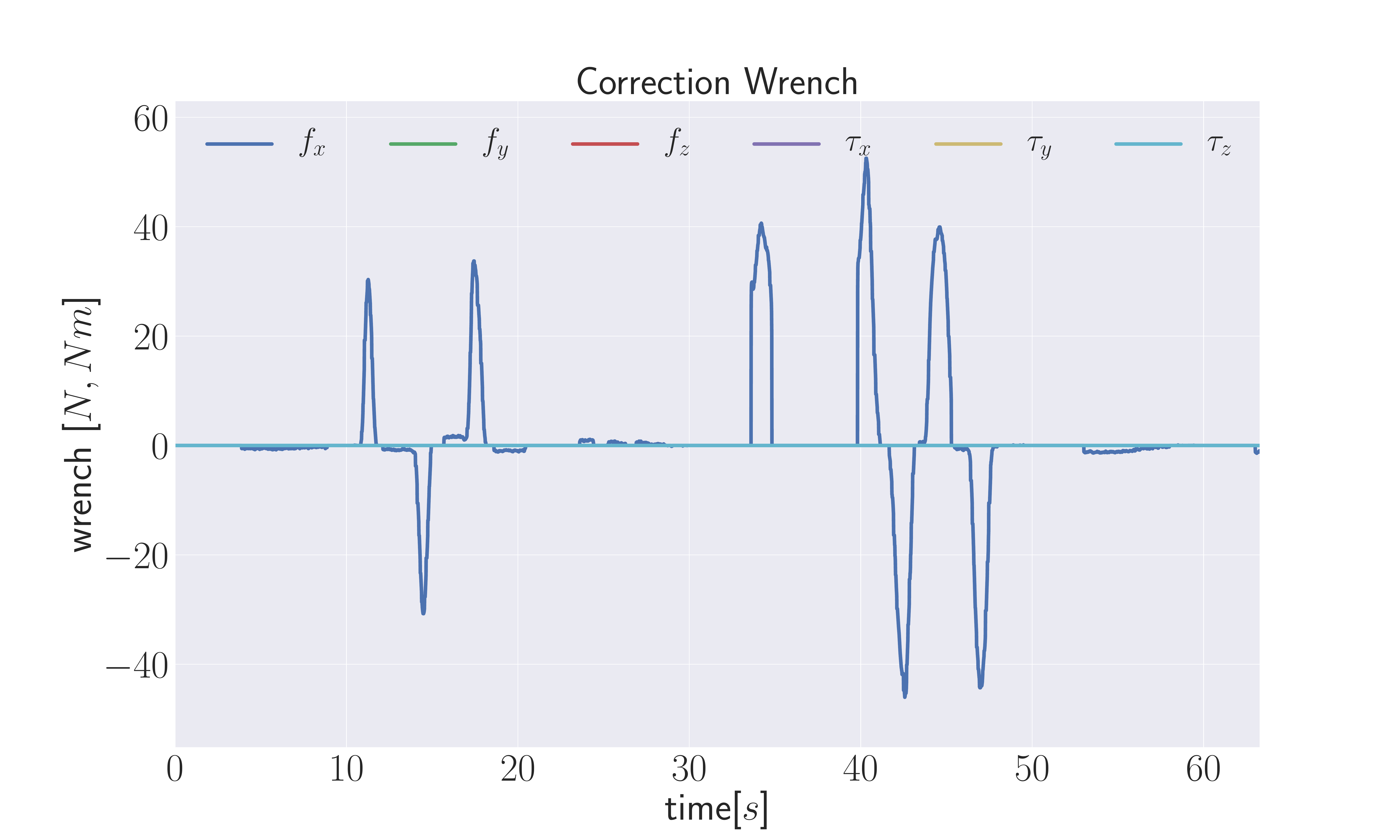}
        \caption{\hspace*{-7.5mm}}
        \label{fig:real-robot-correction-wrench-x}
    \end{subfigure}
    \begin{subfigure}{0.5\textwidth}
        \centering
        \includegraphics[clip, trim=1cm 0.5cm 4.5cm 2.5cm, scale=0.125]{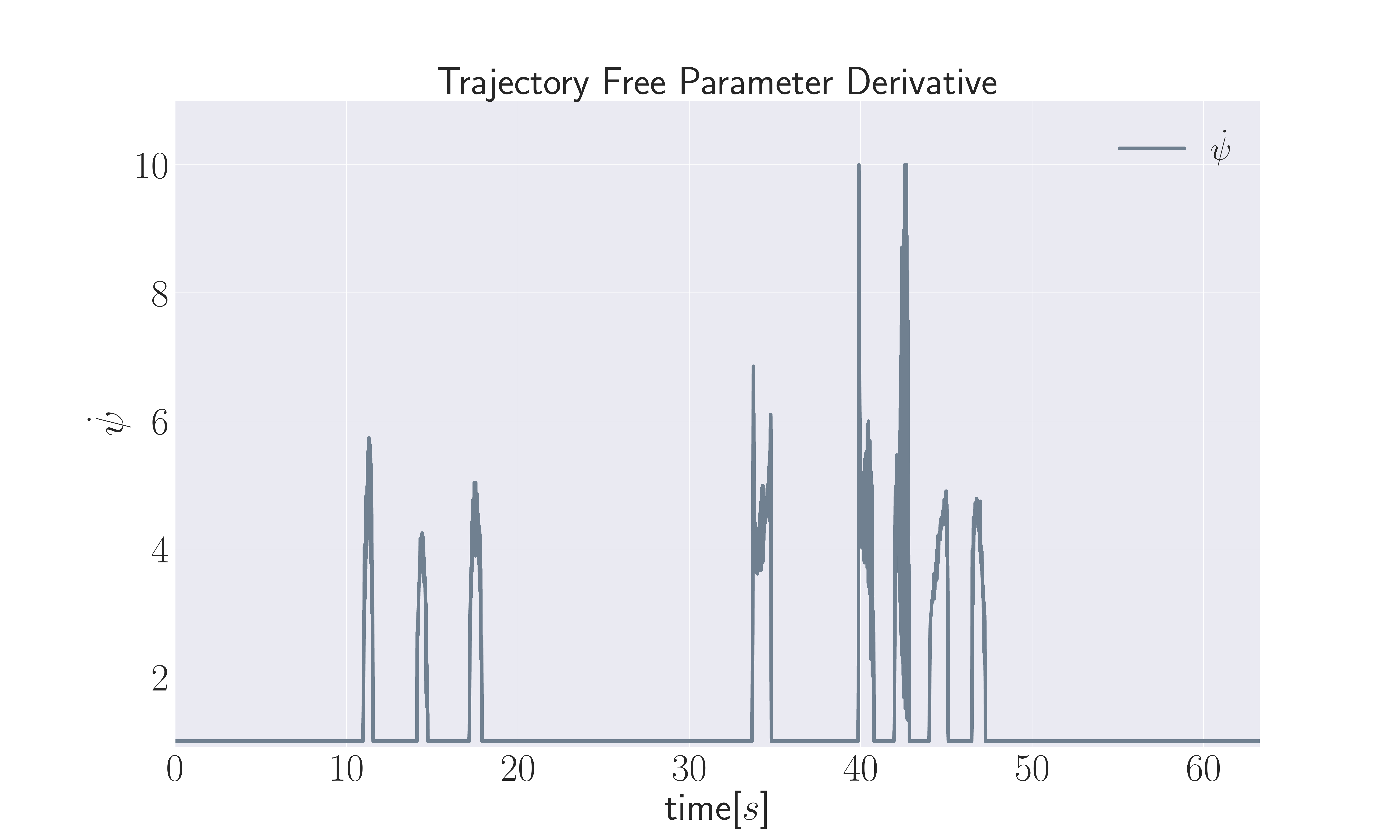}
        \caption{\hspace*{-7.5mm}}
        \label{fig:real-robot-sdotvalue-x}
    \end{subfigure}%
    \begin{subfigure}{0.5\textwidth}
        \centering
        \includegraphics[clip, trim=1cm 0.5cm 4.5cm 2.5cm, scale=0.125]{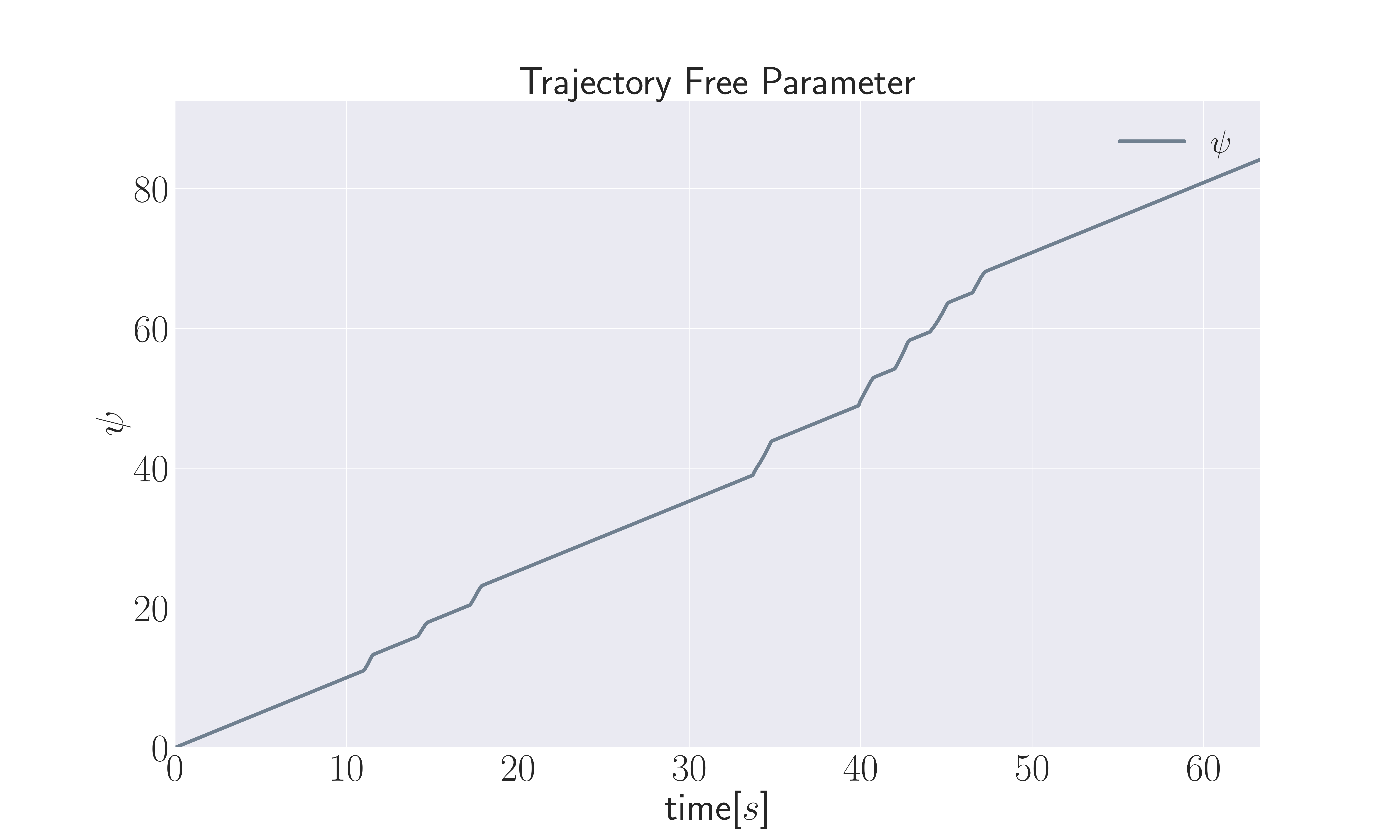}
        \caption{\hspace*{-7.5mm}}
        \label{fig:real-robot-svalue-x}
    \end{subfigure}
    \begin{subfigure}{0.5\textwidth}
        \centering
        \includegraphics[clip, trim=1cm 0.5cm 4.5cm 2.5cm, scale=0.125]{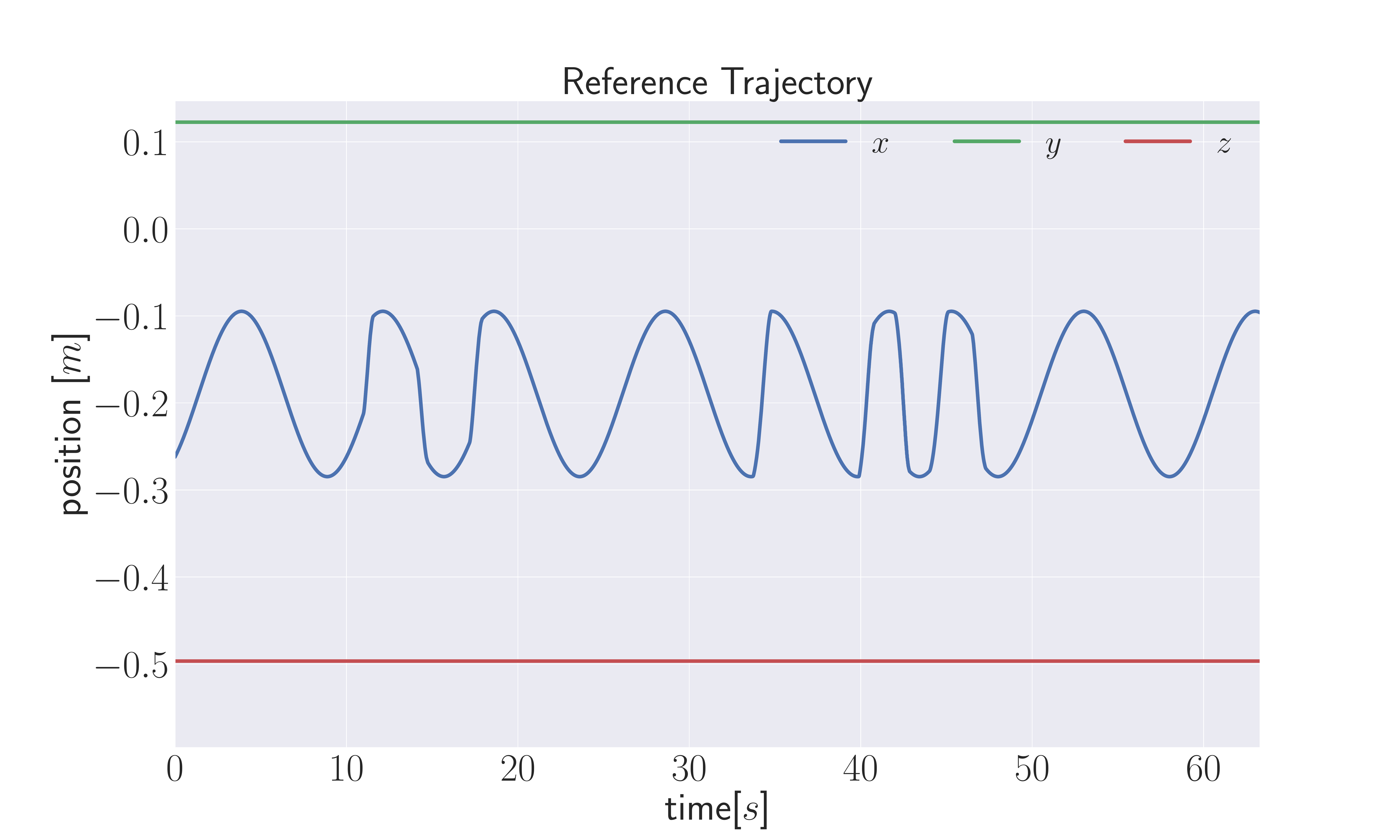}
        \caption{\hspace*{-7.5mm}}
        \label{fig:real-robot-reference-trajectory-x}
    \end{subfigure}%
    \begin{subfigure}{0.5\textwidth}
        \centering
        \includegraphics[clip, trim=1cm 0.5cm 4.5cm 2.5cm, scale=0.125]{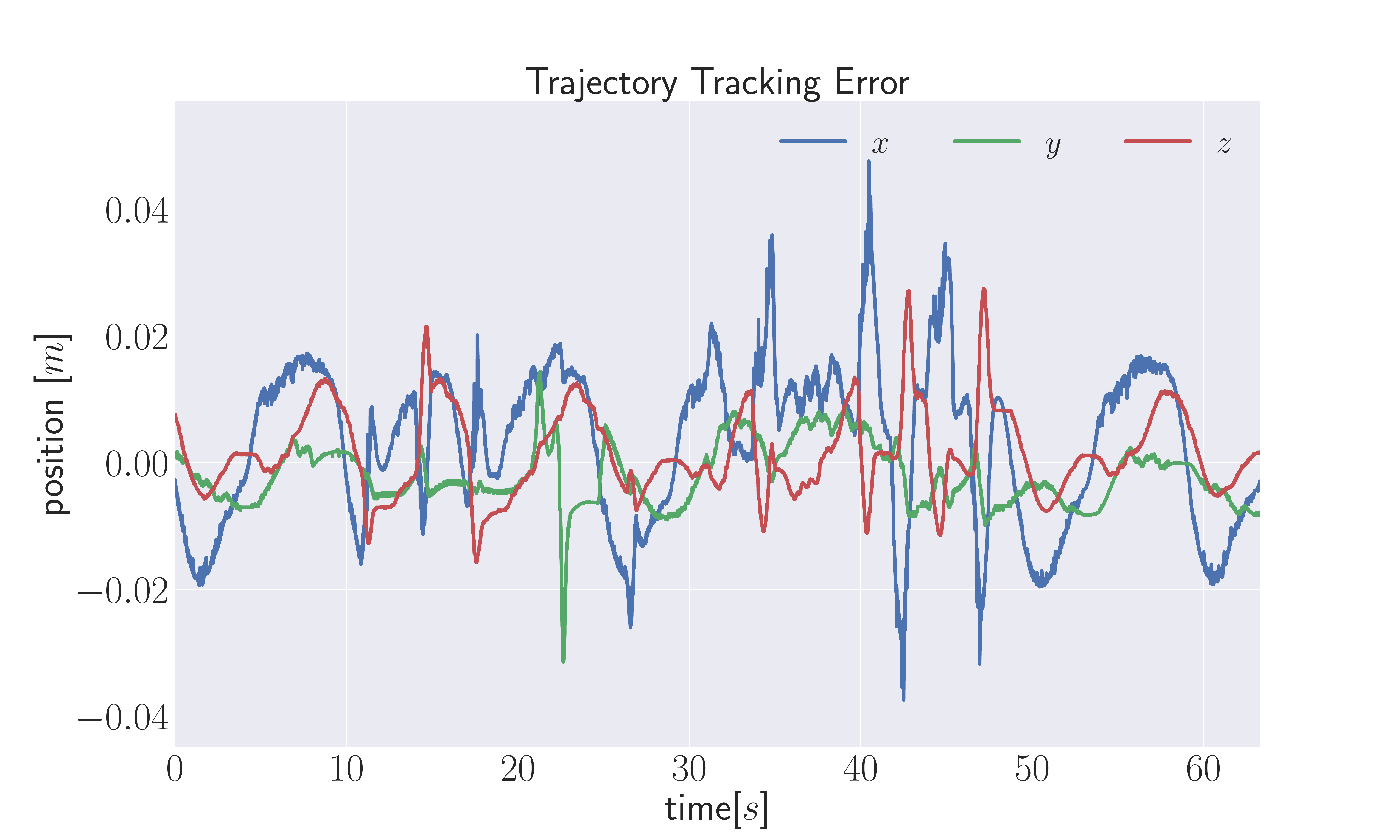}
        \caption{\hspace*{-7.5mm}}
        \label{fig:real-robot-trajectory-tracking-error-x}
    \end{subfigure}
    \caption{Real robot 1D trajectory advancement along $X$-direction}
    \label{fig:real-robot-velocity-modulation-1d-x}
\end{figure*}

The results of experiments on the real icub robot with 1D reference trajectory along the x-axis is shown in Fig.~\ref{fig:real-robot-velocity-modulation-1d-x}. The external interaction wrenches experienced by the right foot of the robot are highlighted in Fig.~\ref{fig:real-robot-external-interaction-wrench-x} and the \textit{correction wrench} that is considered towards trajectory advancement is shown in Fig.~\ref{fig:real-robot-correction-wrench-x}. The reference trajectory is similar to a time parametrized trajectory i.e., $\psi = t$ until any helpful wrench is applied to the end-effector. Under the influence of assistive wrenches, the derivative of the trajectory free parameter $\dot{\psi}$ changes as shown in Fig.~\ref{fig:real-robot-sdotvalue-x} and the corresponding trajectory advancement is reflected as an increase in $\psi$ as seen in Fig.~\ref{fig:real-robot-svalue-x}. Accordingly, the reference is advanced further along the reference trajectory as shown in Fig.~\ref{fig:real-robot-reference-trajectory-x}. Furthermore, starting from $t = 30 \si{\second}$ wrench is applied in the positive $x$-direction continuously. While the reference trajectory is in the positive $x$-direction, this wrench is considered assistive but as the reference trajectory is changed to the negative $x$-direction the wrench becomes agnostic and the reference trajectory is unchanged. Although there are some noisy wrenches that are considered to be correction wrench, they are tuned out by a regularization parameter in computing $\dot{\psi}$ to not have any direct effect on trajectory advancement.

The trajectory tracking error on the real robot is highlighted in Fig.~\ref{fig:real-robot-trajectory-tracking-error-x}. Although the tracking error is higher due to phase delays induced by joint friction, the desired amplitude of the reference trajectory is reached.

\section{CONCLUSIONS}
\label{conclusions}

Physical interactions play a crucial role in cooperative and collaborative tasks between humans and robots. In such scenarios, it is desirable to endow robotic systems with the capabilities to make use of any physical interactions towards successful task completion. Our trajectory advancement approach facilitates advancing along a reference trajectory by leveraging assistance from helpful interaction wrench present during human-robot collaboration scenarios. We validated our approach through experiments conducted with the iCub humanoid robot both in simulation and on the real robot. 

Although the tasks demonstrated in this paper are simple, our approach is equally applicable to complicated HRC scenarios such as a complex humanoid robot leveraging assistance to perform a sit-to-stand transition while tracking the center of mass trajectory with momentum control as the primary control objective for maintaining its stability.


\bibliographystyle{IEEEtran}
\bibliography{references}

\end{document}